\documentclass[]{style}

\usepackage[toc,page,header]{appendix}
\usepackage{enumitem}
\usepackage{graphicx}
\usepackage{amsmath,amssymb}
\usepackage{multirow}
\usepackage{bm}
\usepackage{xspace}
\usepackage{ragged2e}
\usepackage{algorithm}
\usepackage{algpseudocode}
\usepackage{booktabs}
\usepackage{colortbl}
\usepackage{orcidlink}
\usepackage{hyperref}
\usepackage{url}
\usepackage{float}

\usepackage{natbib}

\usepackage{CJKutf8}

\usepackage{xargs}  

\usepackage{todonotes}  

\usepackage{multirow}

\usepackage{cleveref}

\usepackage{amsmath}
\usepackage{dsfont}

\usepackage{svg}

\usepackage{mathrsfs}
\usepackage{adjustbox}
\usepackage{multirow}
\usepackage{multicol}
\usepackage{tcolorbox}
\usepackage{changepage}
\usepackage{enumitem}
\usepackage{graphicx}
\usepackage{amssymb}
\usepackage{xcolor}
\usepackage{float}
\usepackage{multirow}
\usepackage{threeparttable}
\usepackage{graphicx}
\usepackage{algorithm}
\usepackage{algpseudocode}
\usepackage{wrapfig}
\usepackage[table]{xcolor}  
\usepackage{colortbl}       
\usepackage{tabularx} 
\usepackage{makecell} 
\usepackage[dvipsnames]{xcolor}

\newcolumntype{g}{>{\columncolor{gray!10}}c} 

\definecolor{catgray}{gray}{0.9}
\definecolor{skyblue}{rgb}{0.53,0.81,0.92} 

\colorlet{skyblue!30}{skyblue!30!white} 

\definecolor{customblue}{RGB}{70,130,180}  

\newtcolorbox{evolbox}[2][]{%
  enhanced,
  colframe=customblue,
  colback=white,
  coltitle=white,
  rounded corners,
  boxrule=1pt,
  titlerule=0pt,
  toptitle=1mm,
  bottomtitle=1mm,
  fonttitle=\bfseries,
  width=#2\textwidth, 
  #1
}

\usepackage{url}

\PassOptionsToPackage{table,xcdraw}{xcolor}
\usepackage{placeins}
\usepackage{pifont}

\definecolor{RowBlue}{HTML}{E9F2FB}
\definecolor{RowRed}{HTML}{F9EAEA}
\definecolor{Top1}{HTML}{50DB4B} 
\definecolor{Top2}{HTML}{A5FFA2} 
\definecolor{Top3}{HTML}{D9FFD9} 
\definecolor{Sub1}{HTML}{EAB8B8}
\definecolor{Sub2}{HTML}{E4E4E4}

\renewcommand{\emph}[1]{\textit{#1}}

\usepackage[dvipsnames]{xcolor}

\usepackage{amsmath,amsfonts,bm}

\def\eqref#1{equation~\ref{#1}}

\def\1{\bm{1}}

\DeclareMathAlphabet{\mathsfit}{\encodingdefault}{\sfdefault}{m}{sl}
\SetMathAlphabet{\mathsfit}{bold}{\encodingdefault}{\sfdefault}{bx}{n}

\definecolor{ourred}{HTML}{3B4F70}
\definecolor{citeblue}{RGB}{0,102,204}
\definecolor{url}{RGB}{0,102,204}
\newcommand{\revisiontext}[1]{#1}

\fancypagestyle{firststyle}{%
  \fancyhf{}
  \fancyfoot[C]{\thepage}%
}
\setlist[itemize]{noitemsep,leftmargin=*,topsep=0em}
\setlist[enumerate]{noitemsep,leftmargin=*,topsep=0em}
\crefname{section}{Sec.}{Secs.}
\Crefname{section}{Section}{Sections}
\crefname{table}{Tab.}{Tabs.}
\Crefname{table}{Table}{Tables}

\title{ModalSearcher: Weaving Visual Evidence into Autonomous Cross-Modal Deep Search}
\author[1,\dagger]{Zhihan~Yin}
\author[2,\dagger]{Ruyi~Li}
\author[3]{Chengyu~Fang}
\author[1]{Yifeng~Yao}
\author[1]{Nonghai~Zhang}
\author[4]{Pengfei~Xiong}
\author[4]{Terry~Zhang}
\author[1,*]{Huishuai~Zhang}
\author[1,5,*]{Dongyan~Zhao}
\affiliation[1]{Peking University}
\affiliation[2]{Southwest University}
\affiliation[3]{Tsinghua University}
\affiliation[4]{Independent Researcher}
\affiliation[5]{National Engineering Research Center of New Electronic Publishing Technologies}
\contribution[\dagger]{Equal contribution.\quad $^{*}$Corresponding authors: Huishuai Zhang and Dongyan Zhao.}
\hypersetup{
  pdfauthor={Zhihan Yin, Ruyi Li, Chengyu Fang, Yifeng Yao, Nonghai Zhang, Pengfei Xiong, Terry Zhang, Huishuai Zhang, Dongyan Zhao},
  pdftitle={ModalSearcher: Weaving Visual Evidence into Autonomous Cross-Modal Deep Search}
}

\abstract{\sloppy
Static and incomplete parametric knowledge limits the performance of multimodal large language models on novel-knowledge and knowledge-intensive tasks, especially those requiring multi-hop evidence acquisition. Existing multimodal agentic search methods often fix visual evidence at the input or a predefined trajectory position, making it difficult for agents to select search modalities as their information needs evolve and to handle complex real-world queries. We define this capability as \emph{autonomous cross-modal search}: an agent chooses among direct reasoning, text-to-text search, text-to-image search, and reverse-image search, actively introduces visual evidence when appropriate, and switches between and integrates modalities across multiple hops. We propose \revisiontext{ModalSearch-Pipeline}, a scalable procedure for constructing an interleaved multimodal knowledge graph. It samples paths with diverse depths and modality compositions and \revisiontext{uses backward question construction to obscure entity names and image targets that would expose reasoning shortcuts while retaining necessary disambiguating clues, yielding} text-only, image-initialized, and arbitrarily interleaved multi-hop problems. We train the ModalSearcher Agent with supervised fine-tuning and reinforcement learning in a live Web environment, and construct ModalSearch-Bench. \revisiontext{Experiments show that ModalSearcher-8B improves average accuracy by 16.2 points over Qwen3-VL-8B under the same agentic workflow across five multimodal knowledge-intensive QA benchmarks, with consistent gains on independently constructed active visual-search benchmarks.}
}

\renewcommand{\checkdataformat}[2][]{{\abstractinfont\noindent \textbf{#1:} #2}}
\checkdata[GitHub]{\url{https://github.com/Hans-M-Yin/ModalSearcher}}

\begin{document}
\maketitle

\section{Introduction}
\label{sec:introduction}

Multimodal large language models (MLLMs) have made rapid progress on visual understanding, reasoning, and multimodal question answering~\citep{liu2023visualinstructiontuning,yin2026mirareinforcingmultimodalreasoning}. However, their parametric knowledge is incomplete and becomes unreliable for knowledge-intensive or time-sensitive questions. Retrieval-augmented generation (RAG) methods address this limitation by introducing external documents or multimodal knowledge bases, but they typically rely on a separately maintained index and a fixed retrieval--generation procedure~\citep{pmlr-v162-borgeaud22a,izacard-grave-2021-leveraging,lewis2020retrieval}. More recent agentic search systems interleave web search with reasoning, allowing a language model to decide when to retrieve evidence and how to use it to support subsequent reasoning and answer generation~\citep{asai2023selfraglearningretrievegenerate}.

Recent work has extended this paradigm to multimodal deep search agents and has demonstrated strong performance on multimodal knowledge-intensive QA~\citep{mmsearchr1,vision-deepresearch,opensearchvl,mmdeepresearch,webwatcher}. Nevertheless, the multimodal search problem is often formulated as visual understanding followed by text-centric knowledge search: the input image is interpreted at the beginning, after which the remaining reasoning chain is primarily resolved in the text space. Some recent methods introduce visual search, but the visual evidence is frequently assigned to a fixed position in the trajectory or explicitly indicated by the question~\citep{visual-seeker}. These settings underrepresent the multi-step cross-modal planning required in real-world information-seeking tasks, where an agent must organize interleaved textual and visual searches, decide when to switch modalities, and synthesize heterogeneous evidence over multiple steps to reach a well-supported answer to the original user query.

We identify this missing capability as \emph{autonomous cross-modal search}. Given a question and the evidence collected so far, an agent should maintain its information need and decide whether to reason directly, perform text-to-text search, retrieve an image through text-to-image search, or perform reverse-image search. The decision should depend on the expected usefulness of each modality: textual search may expose entities or relations, whereas visual search may reveal clues absent from textual records. The agent must therefore decide not only what to search, but also which modality to use and when to switch modalities, while integrating the resulting evidence over multiple hops.

To address this issue, we introduce \revisiontext{ModalSearch-Pipeline}, a scalable procedure for constructing cross-modal multi-hop search data. It begins with interleaved multimodal graph construction, which alternates text-node and image-node expansion. Text nodes undergo relation-aware textual expansion and retrievability-aware visual expansion; only stably retrievable images receive text-to-image incoming edges, while retrieval-ambiguous images may be retained as initial visual inputs. Image-node expansion uses verified visual entity linking, allowing visual evidence to appear at intermediate positions in a reasoning chain. We then sample paths with diverse depths and modality compositions. Finally, the pipeline merges intermediate text--image--text relations and \revisiontext{applies backward question rewriting to obscure entity names and image targets that would expose reasoning shortcuts, while preserving necessary disambiguating clues and avoiding explicit modality-switch instructions.} The resulting questions support text-only, image-initialized, and arbitrarily interleaved multimodal reasoning while requiring the agent to discover the appropriate evidence through search.

Building on this data, we develop the ModalSearcher Agent, which interacts with a real search environment through text-to-text search, text-to-image search, reverse-image search, and URL reading. At each round, the agent observes retrieved evidence, identifies its remaining information need, and either answers or invokes another tool. We train the agent in two stages: supervised fine-tuning teaches ReAct-style tool use~\citep{yao2023react} and evidence-grounded reasoning, while reinforcement learning further optimizes multi-round search decisions in the environment. The graph paths are used during question construction and expert-trajectory generation, but are not exposed to the agent at inference time. We also construct the ModalSearch-Bench, a curated benchmark designed to stress autonomous modality selection and cross-modal evidence integration.

Experiments show that ModalSearcher-\allowbreak 8B improves over Qwen3-VL-8B by 19.8 and 15.8 percentage points on the independently constructed VisBrowse~\citep{visbrowse} and BrowseComp-V3~\citep{browsecomp-v3} benchmarks, respectively, providing evidence of transfer beyond the construction procedure used by \revisiontext{ModalSearch-Pipeline}. On five established multimodal QA and knowledge-search benchmarks, the model achieves an average score of 57.0, improving over Qwen3-VL-8B by 16.2 points. ModalSearch-Bench complements these external evaluations with a controlled assessment under the same construction paradigm as the training data. Ablations of the data pipeline, search tools, and training stages support the value of interleaved modality paths and modality-aware tool decisions for the observed gains. Our contributions are summarized as follows:

\begin{itemize}
    \item We introduce \revisiontext{ModalSearch-Pipeline} and ModalSearch-Bench. The pipeline constructs an interleaved multimodal graph, samples modality-diverse reasoning paths, and synthesizes questions with implicit visual dependencies and controllable search depth; the benchmark evaluates autonomous modality selection and cross-modal evidence integration.
    \item We formulate multimodal deep search as autonomous cross-modal search and develop the ModalSearcher Agent, which decides what to search, which modality to use based on its evolving information need.
    \item We train the ModalSearcher Agent and use benchmarks and ablations to evaluate the contributions of generated data, multimodal tools, and training.
\end{itemize}

\section{Related Work}
\label{sec:related-work}

\subsection{Agentic Search}

To alleviate the limitations of parametric knowledge and solve knowledge-intensive tasks, early retrieval-augmented generation (RAG) methods relied on a pre-built external knowledge base or document index~\citep{pmlr-v162-borgeaud22a,izacard-grave-2021-leveraging}. An independent retriever searched for relevant information before generation and injected it into the context~\citep{lewis2020retrieval,3524938.3525306}. These methods require maintenance of the knowledge base and retrieval system, and their fixed retrieval--generation procedure is difficult to adapt to open-domain long-horizon tasks. With the development of tool-augmented agents, recent work has explicitly interleaved web search with the reasoning process of large language models, allowing them to issue searches dynamically according to their current reasoning state and integrate external evidence~\citep{asai2023selfraglearningretrievegenerate}. For example, Search-o1 uses agentic RAG to generate search queries when the model encounters knowledge uncertainty and to reason further over retrieved documents~\citep{li-etal-2025-search}. R1-Searcher and Search-R1 use outcome-based reinforcement learning to train models to learn search invocation, query generation, and search-result utilization~\citep{song2025r1searcherincentivizingsearchcapability,jin2025searchr}. DeepResearcher, WebThinker, and WebSailor extend training or inference to the real open Web and study multi-round search, browsing, and evidence integration under noisy, redundant, and incomplete webpage observations~\citep{zheng-etal-2025-deepresearcher,li2026webthinker,li2025websailornavigatingsuperhumanreasoning}. MiroThinker further explores sustained interaction in long-horizon deep-research agents through context management and interactive scaling~\citep{miromindteam2026mirothinkerpushingperformanceboundaries}. Together, these studies show that interleaving search behavior with reasoning can improve factuality and reliability on knowledge-intensive, multi-hop, and open-domain long-horizon tasks while mitigating hallucinations caused by missing knowledge~\citep{yu2026browseragent,yin2026freakfinegrainedhallucinationevaluation}.

\subsection{Multimodal Deep Search Agents}

Inspired by text-only search agents, recent work has extended agentic search to multimodal large language models and improved performance on knowledge-intensive visual QA~\citep{webwatcher,mmdeepresearch-r1,SenseNova-MARS,DeepEyes-V2}. MMSearch-R1~\citep{mmsearchr1} trains MLLMs to invoke text-to-image and text-to-text search tools through reinforcement learning, while Vision-DeepResearch~\citep{vision-deepresearch}, OpenSearch-VL~\citep{opensearchvl}, and MM-DeepResearch~\citep{mmdeepresearch} construct multimodal search trajectories for multi-hop reasoning. Other methods use cached search results or offline sandboxes to improve training efficiency and stability~\citep{searcheyes}. Despite this progress, many existing formulations remain largely text-centric: visual search is primarily used to recognize entities or local regions in the input image, while subsequent multi-hop retrieval and reasoning are conducted in the text space~\citep{visual-seeker,mta-agent,ode}. \revisiontext{VisBrowse and BrowseComp-V3 evaluate active image retrieval, sharing our goal of acquiring new visual information during reasoning~\citep{visbrowse,browsecomp-v3}. Our training data support task-driven visual search interleaved with textual retrieval.} Visual-Seeker~\citep{visual-seeker} also uses visual search, but its visual target is usually fixed and may be explicitly indicated in the question. \revisiontext{InterLV-Search~\citep{interLV} evaluates interleaved multimodal search in which retrieved visual evidence guides subsequent retrieval. Our work focuses on graph-based supervision synthesis and agent training, complemented by ModalSearch-Bench.} We study task-driven autonomous cross-modal search, where the agent determines the search modality and target from its evolving information need~\citep{zhang2026deepvoyagervlincentivizingvisionintheloopsearch}. Our formulation jointly addresses modality selection, modality switching, and cross-modal evidence integration, and is supported by data that cover different search depths and interleaved textual and visual reasoning.

\begin{figure}[t]
    \centering
    \caption{Comparison of multimodal search task formulations: input-image understanding, explicitly specified visual-search targets, and autonomous acquisition of visual evidence during reasoning.}
    \label{fig:method_compare}
    \includegraphics[width=\linewidth]{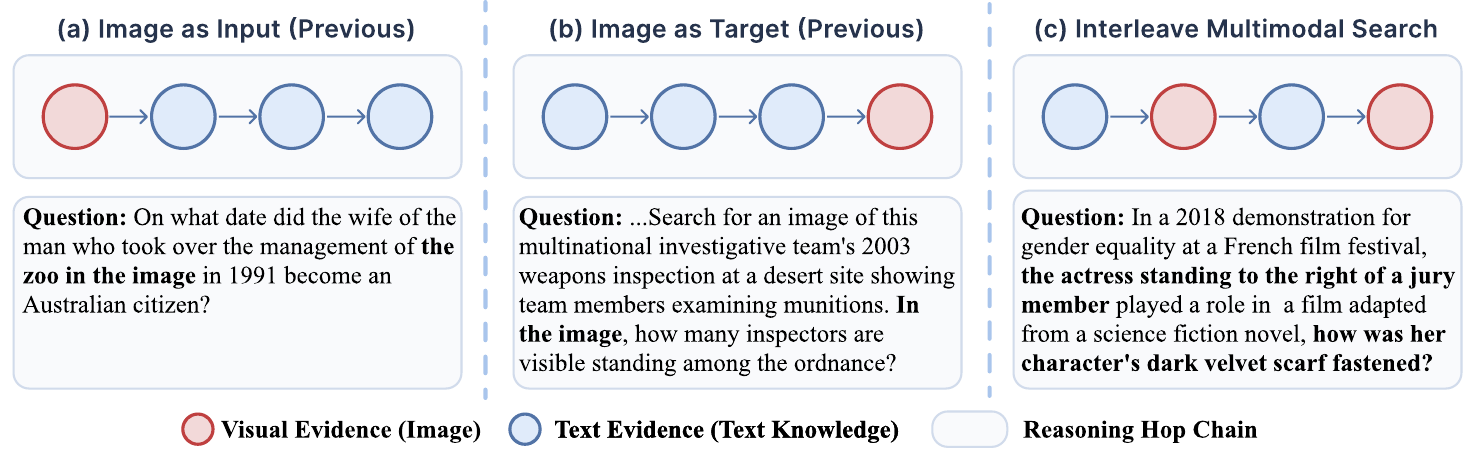}
\end{figure}

\section{Method}
\label{sec:method}

\revisiontext{Our method centers on ModalSearch-Pipeline, a scalable data-synthesis pipeline for autonomous cross-modal search.} The pipeline comprises three stages. First, interleaved multimodal graph construction combines two text-node operations, namely relation-aware textual expansion and retrievability-aware visual expansion, with image-node expansion through verified visual entity linking. Second, we sample reasoning paths with diverse depths and modality compositions. Third, question synthesis combines edge descriptions along text--image--text subpaths and \revisiontext{applies backward question rewriting to obscure entity names and visual targets that would expose reasoning shortcuts, while preserving necessary disambiguating clues and avoiding explicit modality-switch instructions.} The resulting data are used to train the ModalSearcher Agent with supervised fine-tuning and reinforcement learning in a live web environment. Importantly, the graph serves only as an offline scaffold for data synthesis and trajectory generation; it's not exposed to the agent as a retrieval index at inference time.

\begin{figure}[t]
    \centering
    \caption{Overview of the proposed data-synthesis pipeline. The pipeline constructs an interleaved multimodal knowledge graph, samples modality-diverse reasoning paths, and \revisiontext{synthesizes questions that obscure shortcuts and visual targets while retaining necessary disambiguating clues and avoiding explicit modality-switch instructions.}}
    \label{fig:data_pipeline}
    \includegraphics[width=\linewidth]{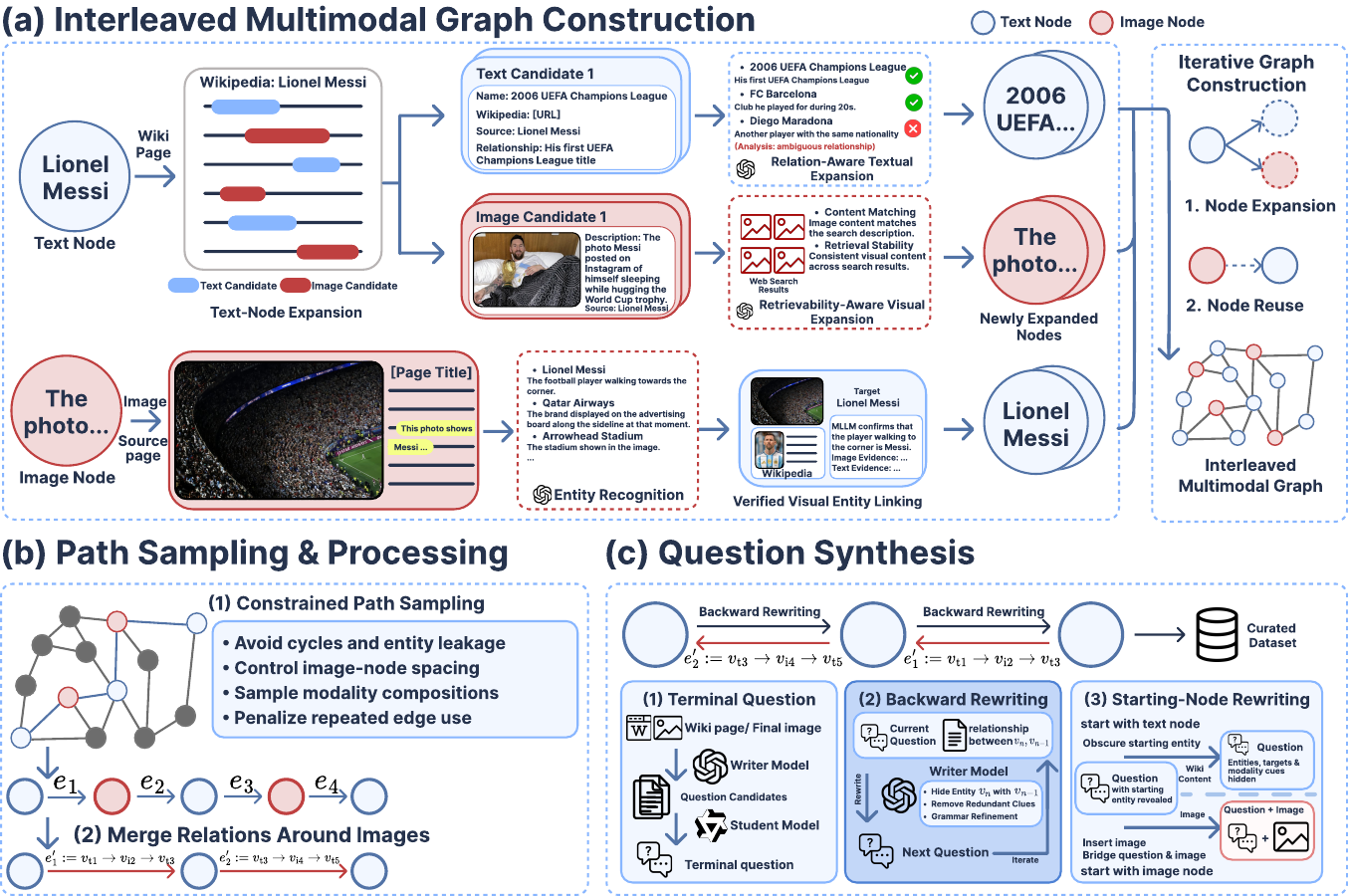}
\end{figure}

\subsection{Interleaved Multimodal Graph Construction}
\label{sec:multimodal-kg}

We represent the graph as $G=(V,E)$, where $V=V_{\mathrm{t}}\cup V_{\mathrm{i}}$ consists of text nodes $V_{\mathrm{t}}$ and image nodes $V_{\mathrm{i}}$. For an entity $a$, let $w(a)$ denote its Wikipedia article; the corresponding text node is $v_{\mathrm{t}}(a)=(a,w(a))$. An image node is represented as $v_{\mathrm{i}}=(x_i,d_i,c_i)$, where $x_i$ is the image content, $d_i$ is its textual description, and $c_i$ is source-webpage metadata. We partition the directed edges into three modality-based categories: text-to-text, text-to-image, and image-to-text, giving $E=E_{\mathrm{t}\rightarrow\mathrm{t}}\cup E_{\mathrm{t}\rightarrow\mathrm{i}}\cup E_{\mathrm{i}\rightarrow\mathrm{t}}$. Each directed edge $e=(v_s,r,v_d)$ is annotated with a natural-language relation description $r$, which expresses the semantic or visual association between the source and destination nodes and provides a clue for subsequent question synthesis.

Starting from multiple seed nodes, graph construction alternates between text-node expansion and image-node expansion. Before a node is scheduled for expansion, we check whether it already exists in the graph. Existing nodes are reused, while unseen nodes are added to the expansion queue; once a proposed neighbor and its relation are validated, a directed edge is inserted between the source and destination nodes. This iterative procedure connects locally proposed neighbors into a graph rather than producing isolated reasoning chains.

Text-node expansion follows two complementary branches. In relation-aware textual expansion, we extract the local hyperlink structure from $w(a)$ to construct neighboring entities $\hat a$, forming the candidate set $\mathcal{C}_{\mathrm{t}}(a)=\{v_{\mathrm{t}}(\hat a)\}$. For each candidate, an MLLM (see Table~\ref{tab:data-generation-config}) uses the local textual context to identify the linked entity and generate the semantic relation between $a$ and $\hat a$. Candidates with weak semantic associations, ambiguous relations, or non-entity targets are removed. For every retained candidate, we construct or reuse $v_{\mathrm{t}}(\hat a)$ and add $e=(v_{\mathrm{t}}(a),r,v_{\mathrm{t}}(\hat a))\in E_{\mathrm{t}\rightarrow\mathrm{t}}$. Thus, neighboring nodes and their relations are proposed and validated jointly. In retrievability-aware visual expansion, an MLLM identifies paragraphs in $w(a)$ that describe visual content retrievable through text, such as a scene, event, or artwork, and converts them into candidate image descriptions or search queries. According to the visual and referential consistency of the returned results, we partition image nodes into the stably retrievable set $V_{\mathrm{i}}^{\mathrm{stable}}$ and the retrieval-ambiguous set $V_{\mathrm{i}}^{\mathrm{amb}}$, where $V_{\mathrm{i}}=V_{\mathrm{i}}^{\mathrm{stable}}\mathbin{\dot\cup}V_{\mathrm{i}}^{\mathrm{amb}}$. If the results for $d_i$ refer to the same visual content, we add $e=(v_{\mathrm{t}}(a),r,v_{\mathrm{i}})\in E_{\mathrm{t}\rightarrow\mathrm{i}}$. Retrieval-ambiguous images receive no text-to-image incoming edge because their descriptions may lead to substantially different yet plausible results, but they may still be retained as initial visual nodes. This asymmetric connectivity separates whether an image is useful in the graph from whether it can be reached reliably from a textual description.

Image-node expansion proceeds through verified visual entity linking. Given $v_{\mathrm{i}}=(x_i,d_i,c_i)$, an MLLM proposes entities appearing in the image using its visual content, description, and source-page context. Each candidate entity $a'$ is linked to a corresponding Wikipedia article and further verified using the article body and images. If $v_{\mathrm{t}}(a')$ already exists, we reuse it; otherwise, we add it to $V_{\mathrm{t}}$ and schedule it for subsequent expansion. We then construct $e=(v_{\mathrm{i}},r,v_{\mathrm{t}}(a'))\in E_{\mathrm{i}\rightarrow\mathrm{t}}$, where $r$ describes the visual or semantic relation between the entity and the image, such as its spatial position, event participation, scene membership, or relation to an artwork.

Alternating these two forms of node expansion, together with modality-typed edge construction and node reuse, yields an interleaved graph that supports text-to-text, text-to-image, and image-to-text transitions over multiple hops. In particular, image nodes are not restricted to initial inputs or terminal evidence: verified image-to-text links allow them to appear at intermediate positions and support paths with arbitrary text--image interleaving. Further details on retrievability assessment, visual entity linking, and graph assembly are provided in Appendix~\ref{app:data-synthesis}.

\subsection{Path Sampling and Question Synthesis}
\label{sec:path-question}

After constructing the graph, we sample reasoning paths with controllable depths and modality compositions. We randomly sample an initial node $v_0\in V$ and a target depth $L\in\{K_{\min},\ldots,K_{\max}\}$, and then perform a constrained random walk along outgoing edges to obtain $\pi=(v_0,v_1,\ldots,v_L)$. The sampler avoids cycles, semantic repetition, overused edges, and relations that prematurely expose later entities, while controlling the spacing of image nodes. These constraints produce diverse text-only, image-initialized, and interleaved multimodal paths rather than merely following the most frequent graph relations.

Before writing a question, we process intermediate text--image--text subpaths. For $v_{\mathrm{t}}(a)\rightarrow v_{\mathrm{i}}\rightarrow v_{\mathrm{t}}(a')$, the image represents a specific scene, event, or artwork associated with $a$, while $v_{\mathrm{t}}(a')$ represents an entity appearing in that image. An LLM combines the two edge descriptions into a composed relation description $\tilde r$, represented as $v_{\mathrm{t}}(a)\xrightarrow{\tilde r}v_{\mathrm{t}}(a')$ for question synthesis, while retaining the original image-mediated subpath as the supporting evidence. The composed description refers to the relevant scene, event, or artwork without explicitly mentioning the image node or instructing the agent to perform text-to-image search. This rewriting is intended to make the visual evidence dependency implicit and encourage autonomous modality selection; it does not by itself rule out alternative textual solution paths.

We next generate a terminal question $q_L$ according to the modality of $v_L$. If $v_L=v_{\mathrm{t}}(a)$, an LLM generates a knowledge question from $w(a)$; if $v_L=v_{\mathrm{i}}$, an MLLM generates a question about visual attributes or relations in the image. If the starting node $v_0$ is an image, its content $x_0$ is provided as the initial visual input. All other image nodes are withheld and must be retrieved by the agent according to its evolving information need.

Finally, backward question rewriting transforms $q_L$ recursively from the end of the path toward its start. For each edge $v_{n-1}\xrightarrow{r_n}v_n$, $n=L,\ldots,1$, the predecessor and relation are used to replace the explicit reference to $v_n$ in $q_n$ with an indirect description, yielding $q_{n-1}$. Redundant or path-irrelevant content is removed, while clues necessary for solvability and answer uniqueness are retained. For a textual starting node $v_0=v_{\mathrm{t}}(a)$, an MLLM further replaces $a$ with an indirect description derived from $w(a)$; for an image starting node, its description is replaced by ``this image.'' \revisiontext{The final question $\tilde q_0$ obscures entity names and image-search targets that would enable search shortcuts and avoids explicit modality-switch instructions.} Together, these steps preserve the need for the agent to determine what evidence is missing, which modality is most informative, and when to switch modalities.

\paragraph{\normalfont\textbf{ModalSearch-Bench construction.}} To obtain a held-out evaluation set, we independently construct a new 20,035-node multimodal knowledge graph with the same ModalSearch-Pipeline procedure and a disjoint set of initial seed nodes, including seeds from domains not covered by the training graph. We apply the same path-sampling and backward question-rewriting steps to this graph and sample 900 candidate questions covering different reasoning depths and modality-path types. Experts retain 300 questions after individually verifying their answers, correctness, and solvability against Wikipedia and removing ambiguous cases. We distinguish Input-only questions, whose reference visual evidence is provided in the initial input, from Search-required questions, which call for actively searching for new visual information during reasoning to advance the solution. The benchmark contains 100 Input-only and 200 Search-required questions, with broadly balanced reasoning depths. Its graph nodes, paths, and questions are kept disjoint from all training data and used exclusively to evaluate autonomous cross-modal search, including modality selection, modality switching, and cross-modal evidence integration; further details are provided in Appendix~\ref{app:benchmark-details}.

\subsection{Agentic Search and Training}
\label{sec:agent-config}
\label{sec:training}

The ModalSearcher Agent uses a ReAct-based interaction framework over a live web environment. Unlike the offline graph transitions, the runtime tools operate on the open web and are selected autonomously based on the agent's information need. At each round, the agent either answers directly or invokes one of four tools:
\begin{itemize}
    \item \texttt{t2t\_search(query)} performs text-to-text search and returns the top $K$ results;
    \item \texttt{t2i\_search(query)} performs text-to-image search and returns the top $K$ images;
    \item \texttt{i2i\_search(image, region)} reverse-searches the cropped region for $K$ related images;
    \item \texttt{read\_url(url)} reads webpages/PDFs as Markdown or downloads images.
\end{itemize}
The search tools return titles, snippets, image information, and URLs. The agent assesses relevance, reads selected sources, and incrementally integrates evidence.

\paragraph{\normalfont\textbf{Supervised fine-tuning.}} We sample questions and their corresponding reasoning paths from the interleaved multimodal graph. For each sample, the question, ground-truth answer, and path transitions $v_{n-1}\xrightarrow{r_n}v_n$ are provided to an expert model as offline auxiliary information. Using only the same tools available to the target agent, the expert interacts with the real search environment and generates ReAct-formatted trajectories. We discard trajectories with incorrect answers, failed tool calls, or insufficient evidence and use the retained samples for SFT. In our experiments, GPT-5.4 generates the expert trajectories. The auxiliary graph paths are never provided to the target model at inference time.

\paragraph{\normalfont\textbf{Reinforcement learning.}} Starting from the SFT model, we use RLOO~\citep{ahmadian-etal-2024-back} to optimize the agent's multi-round search decisions in the real web environment. For each question, we sample a group of $K$ trajectories $\{\tau_1,\ldots,\tau_K\}$. Let $r(\tau_i)$ denote the trajectory-level reward, whose design is described below. RLOO uses the average reward of the other trajectories in the same group as a leave-one-out baseline:
\[
b_i=\frac{1}{K-1}\sum_{\substack{j=1\\j\neq i}}^{K}r(\tau_j),
\qquad
A_i=r(\tau_i)-b_i,
\]
where $A_i$ is held fixed during the policy update. Let $\mathcal{B}$ denote a batch of trajectories, with advantages computed within their question groups. For token $y_{i,t}$ in a trajectory of token-sequence length $T_i$, $h_{i,t}$ contains the question and preceding multimodal interaction history, including tool observations. The mask $m_{i,t}$ is $1$ for model-generated response tokens participating in optimization and $0$ for prompt, tool-observation, and padding tokens. The token-mean policy-gradient surrogate averages over all unmasked response tokens in $\mathcal{B}$:
\[
\mathcal{L}_{\mathrm{RLOO}}(\theta)
=-
\frac{\sum_{i\in\mathcal{B}}\sum_{t=1}^{T_i}
m_{i,t}A_i\log\pi_{\theta}(y_{i,t}\mid h_{i,t})}
{\sum_{i\in\mathcal{B}}\sum_{t=1}^{T_i}m_{i,t}}.
\]
This reduction does not first normalize each trajectory by its individual response length. We define the reward of trajectory $\tau_i$ as
\[
r(\tau_i)=\alpha r_{\mathrm{acc}}(\tau_i)
+\beta r_{\mathrm{fmt}}(\tau_i),
\]
where $r_{\mathrm{acc}}$ is $1$ when the final answer is judged to be correct and $0$ otherwise, and $r_{\mathrm{fmt}}$ is $1$ when all tool calls follow the predefined interaction format and $0$ otherwise. For rollouts terminated by repeated agent-caused format or tool-call errors, we override this reward with $0.1$; tool-instability failures receive $0$, as detailed in Appendix~\ref{app:rl-details}. During training, we cache tool-call results within each group, which improves the stability of web search to some extent and reduces the training cost.

\begin{table}[!t]
    \setlength{\abovecaptionskip}{2pt}
    \setlength{\belowcaptionskip}{2pt}
    \definecolor{tablebestblue}{RGB}{80,100,164}
    \definecolor{tablesecondgreen}{RGB}{108,160,172}
    \caption{
    Performance on multimodal knowledge-intensive QA and web-search benchmarks, including MMSearch~\citep{jiang2024mmsearchbenchmarkingpotentiallarge}, BrowseComp-VL~\citep{geng2025webwatcherbreakingnewfrontier}, VDR-Bench~\citep{zeng2026vision}, FVQA~\citep{wang2017fvqafactbasedvisualquestion}, and LiveVQA~\citep{fu2026seeking}.
    \textcolor{tablebestblue}{\textbf{Bold}} and \textcolor{tablesecondgreen}{\underline{underline}} mark the best and second-best scores in each column \revisiontext{within each group}, respectively.
    }
    \centering
    \small
    \setlength{\tabcolsep}{4.8pt}
    \renewcommand{\arraystretch}{1.2}
    \definecolor{tabledeltagreen}{HTML}{228B22}
    \renewcommand{\textbf}[1]{{\bfseries\sffamily #1}}
    \begin{tabularx}{\textwidth}{@{}>{\raggedright\arraybackslash}p{0.30\textwidth}*{6}{>{\centering\arraybackslash}X}@{}}
    \toprule
    \textbf{Model} & \textbf{VDR} & \textbf{MMSearch} & \textbf{LiveVQA} & \textbf{BC-VL} & \textbf{FVQA} & \textbf{Avg} \\
    \midrule
    \rowcolor{gray!15}
    \multicolumn{7}{c}{\textit{Direct Reasoning}} \\
    GPT-5 \citep{singh2026openaigpt5card} & \textcolor{tablebestblue}{\textbf{9.8}} & \textcolor{tablesecondgreen}{\underline{35.1}} & 44.4 & \textcolor{tablebestblue}{\textbf{48.6}} & \textcolor{tablesecondgreen}{\underline{54.4}} & \textcolor{tablesecondgreen}{\underline{38.5}} \\
    Gemini-2.5-Pro \citep{comanici2025gemini25pushingfrontier} & \textcolor{tablesecondgreen}{\underline{8.0}} & \textcolor{tablebestblue}{\textbf{39.8}} & \textcolor{tablebestblue}{\textbf{60.3}} & \textcolor{tablesecondgreen}{\underline{43.1}} & \textcolor{tablebestblue}{\textbf{60.7}} & \textcolor{tablebestblue}{\textbf{42.4}} \\
    Claude-4-Sonnet & 2.0 & 18.7 & 38.5 & 29.3 & 35.3 & 24.8 \\
    Qwen3-VL-8B \citep{bai2025qwen3vltechnicalreport} & 2.4 & 15.0 & 41.7 & 21.8 & 27.9 & 21.8 \\
    Qwen3-VL-32B \citep{bai2025qwen3vltechnicalreport} & 4.0 & 23.7 & \textcolor{tablesecondgreen}{\underline{49.7}} & 28.3 & 35.6 & 28.3 \\
    \midrule
    \rowcolor{gray!15}
    \multicolumn{7}{c}{\textit{Agentic Workflow}} \\
    GPT-5 \citep{singh2026openaigpt5card} & \textcolor{tablebestblue}{\textbf{22.3}} & 65.7 & \textcolor{tablebestblue}{\textbf{73.3}} & \textcolor{tablebestblue}{\textbf{49.1}} & \textcolor{tablebestblue}{\textbf{69.0}} & \textcolor{tablebestblue}{\textbf{55.9}} \\
    Claude-4-Sonnet & 13.6 & \textcolor{tablebestblue}{\textbf{70.3}} & 69.7 & 44.1 & \textcolor{tablebestblue}{\textbf{69.0}} & 53.3 \\
    Gemini-2.5-Pro \citep{comanici2025gemini25pushingfrontier} & \textcolor{tablesecondgreen}{\underline{18.8}} & \textcolor{tablesecondgreen}{\underline{67.8}} & \textcolor{tablesecondgreen}{\underline{70.0}} & \textcolor{tablesecondgreen}{\underline{45.2}} & \textcolor{tablesecondgreen}{\underline{68.3}} & \textcolor{tablesecondgreen}{\underline{54.0}} \\
    Qwen3-VL-8B \citep{bai2025qwen3vltechnicalreport} & 15.2 & 52.0 & 53.0 & 31.6 & 51.9 & 40.7 \\
    Qwen3-VL-32B \citep{bai2025qwen3vltechnicalreport} & 17.1 & 58.3 & 64.3 & 37.1 & 60.1 & 47.4 \\
    \midrule
    \rowcolor{gray!15}
    \multicolumn{7}{c}{\textit{Deep Search Agent}} \\
    MMSearch-R1-7B \citep{mmsearchr1} & 2.9 & 53.8 & 48.4 & 20.9 & 58.4 & 36.9 \\
    DeepEyes-v2-7B \citep{hong2026deepeyesv2} & 7.8 & 63.7 & -- & -- & 60.6 & -- \\
    WebWatcher-7B \citep{webwatcher} & 10.3 & 49.1 & 51.2 & 21.2 & -- & -- \\
    WebWatcher-32B \citep{webwatcher} & -- & 55.3 & 58.7 & 27.0 & -- & -- \\
    SenseNova-MARS-8B \citep{SenseNova-MARS} & 19.4 & 67.4 & 56.2 & 35.1 & 67.1 & 49.0 \\
    ODE-8B \citep{ode} & 20.4 & 66.0 & -- & \textcolor{tablesecondgreen}{\underline{41.9}} & 64.7 & -- \\
    MM-DeepResearch-8B \citep{mmdeepresearch} & - & \textcolor{tablesecondgreen}{\underline{67.8}} & \textcolor{tablesecondgreen}{\underline{65.0}} & 37.9 & 69.2 & - \\
    OpenSearch-VL-8B \citep{opensearchvl} & \textcolor{tablesecondgreen}{\underline{20.8}} & 64.5 & 59.6 & 37.6 & \textcolor{tablebestblue}{\textbf{71.5}} & \textcolor{tablesecondgreen}{\underline{50.8}} \\
    \midrule
    \textbf{ModalSearcher-8B (Ours)} & \textcolor{tablebestblue}{\textbf{26.2}} & \textcolor{tablebestblue}{\textbf{70.3}} & \textcolor{tablebestblue}{\textbf{70.3}} & \textcolor{tablebestblue}{\textbf{48.1}} & \textcolor{tablesecondgreen}{\underline{69.9}} & \textcolor{tablebestblue}{\textbf{57.0}} \\
    \cellcolor{gray!15}$\Delta$ vs Qwen3-VL-8B & \cellcolor{gray!15}\textcolor{tabledeltagreen}{+11.0} & \cellcolor{gray!15}\textcolor{tabledeltagreen}{+18.3} & \cellcolor{gray!15}\textcolor{tabledeltagreen}{+17.3} & \cellcolor{gray!15}\textcolor{tabledeltagreen}{+16.5} & \cellcolor{gray!15}\textcolor{tabledeltagreen}{+18.0} & \cellcolor{gray!15}\textcolor{tabledeltagreen}{+16.2} \\
    \bottomrule
    \end{tabularx}
    \label{tab:main_results}
\end{table}

\section{Experiments}
\label{sec:experiments}

\subsection{Experiment Setup}
\label{sec:experiment-setup}

\paragraph{\normalfont\textbf{Datasets and evaluation benchmarks.}} We construct an approximately 16K-example SFT dataset and use 5.8K additional pipeline-generated examples for RL, covering diverse reasoning depths, domains, and modality-path types. The SFT set combines 13.5K pipeline-generated examples with 1.3K examples from the FVQA training split and 1.2K from the LiveVQA training split to broaden question-type coverage and reduce the risk of overfitting. GPT-5.4 generates the SFT search trajectories. We evaluate multimodal deep-search ability on MMSearch~\citep{jiang2024mmsearchbenchmarkingpotentiallarge}, BrowseComp-VL~\citep{geng2025webwatcherbreakingnewfrontier}, VDR-Bench~\citep{zeng2026vision}, FVQA~\citep{wang2017fvqafactbasedvisualquestion}, and LiveVQA~\citep{fu2026seeking}, and autonomous cross-modal search on VisBrowse~\citep{visbrowse}, BrowseComp-V3~\citep{browsecomp-v3}, and ModalSearch-Bench. \revisiontext{FVQA is included to assess knowledge-intensive visual question answering.}

\paragraph{\normalfont\textbf{Baseline models.}} We select both closed-source and open-source agentic search models as baselines. The closed-source models include GPT-5~\citep{singh2026openaigpt5card}, Claude-4-Sonnet, and Gemini-2.5-Pro~\citep{comanici2025gemini25pushingfrontier}. The open-source models include Qwen3-VL-8B and Qwen3-VL-32B~\citep{bai2025qwen3vltechnicalreport}, SenseNova-MARS-8B~\citep{SenseNova-MARS}, ODE-8B~\citep{ode}, OpenSearch-VL-8B~\citep{opensearchvl}, MMSearch-R1-7B~\citep{mmsearchr1}, DeepEyes-V2-7B~\citep{hong2026deepeyesv2}, WebWatcher-7B~\citep{webwatcher}, and MM-DeepResearch-8B~\citep{mmdeepresearch}. For general models without native search capabilities, we evaluate them in the same tool environment.

\paragraph{\normalfont\textbf{Implementation details.}} We use Qwen3-VL-8B~\citep{bai2025qwen3vltechnicalreport} as the base model. The SFT learning rate is set to $2\times10^{-5}$ for 4 epochs, and the RL learning rate is set to $1\times10^{-6}$ for 4 epochs. Unless otherwise specified, we use GPT-4o as the judge model during evaluation and follow the inference sampling configuration recommended by Qwen3-VL~\citep{bai2025qwen3vltechnicalreport}. For evaluation results, we report mean accuracy over five inference repeated runs. Further implementation details are provided in Appendix~\ref{app:implementation-details}.

\subsection{Main Results}
\label{sec:main-results}

\textbf{ModalSearcher improves multimodal knowledge-intensive QA.} Table~\ref{tab:main_results} compares ModalSearcher-8B with baseline methods on several benchmarks that emphasize visual understanding and knowledge-intensive question answering. Under a unified evaluation protocol, we compare direct reasoning, agentic workflows, and trained deep-search agents. At a comparable parameter scale, ModalSearcher-8B achieves leading performance on multiple benchmarks and surpasses closed-source models on some of them. Compared with Qwen3-VL-8B evaluated directly under the same agentic framework and tool environment, ModalSearcher-8B improves average performance by $16.2$ points and substantially outperforms existing open-source deep-search agents. \revisiontext{Compared with OpenSearch-VL~\citep{opensearchvl}, which is trained on data synthesized by its own pipeline, ModalSearcher achieves higher average accuracy on multimodal knowledge-intensive QA benchmarks and consistent gains on the shared active visual-search benchmarks (Tables~\ref{tab:main_results} and~\ref{tab:interleave_benchs}), supporting the effectiveness of ModalSearch-Pipeline.} \revisiontext{On FVQA, ModalSearcher achieves $69.9\%$, compared with $71.5\%$ for OpenSearch-VL.} Further ablations in Sec.~\ref{sec:ablation} show that ModalSearcher-8B trained with SFT alone already matches or exceeds most baselines, indicating that the constructed training data effectively improves search ability.

\begin{table}[!t]
\centering
\small
\setlength{\tabcolsep}{2pt}
\renewcommand{\arraystretch}{1.15}
\definecolor{tablebestblue}{RGB}{80,100,164}
\definecolor{tablesecondgreen}{RGB}{108,160,172}
\renewcommand{\textbf}[1]{{\bfseries\sffamily #1}}
\caption{Performance on autonomous cross-modal search benchmarks, including VisBrowse~\citep{visbrowse} and BrowseComp-V3~\citep{browsecomp-v3}. The first two rows use direct answering; ``w/ tools'' denotes tool-augmented evaluation. IO and SR denote Input-only (100 questions) and Search-required (200 questions). Bold and underlined scores mark the best and second-best open-source results. We report the mean accuracy over 5 runs.}
\label{tab:interleave_benchs}
\begin{tabularx}{\textwidth}{@{}>{\raggedright\arraybackslash}p{0.25\textwidth}*{5}{>{\centering\arraybackslash}X}c@{}}
\toprule
\multirow{2}{*}{\textbf{Model}} & \multirow{2}{*}{\textbf{VisBrowse}} & \multirow{2}{=}{\centering\shortstack{\textbf{BrowseComp}\\\textbf{-V3}}} & \multicolumn{3}{c}{\textbf{ModalSearch}} & \multirow{2}{*}{\textbf{Avg.}} \\
\cmidrule(lr){4-6}
& & & \textbf{IO} & \textbf{SR} & \textbf{Avg.} & \\
\midrule
GPT-5~\citep{singh2026openaigpt5card} & 30.2 & 13.8 & 66.0 & 37.8 & 47.2 & 30.4 \\
Gemini-2.5-Pro~\citep{comanici2025gemini25pushingfrontier} & 23.4 & 13.5 & 53.5 & 29.5 & 37.5 & 24.8 \\
\midrule
GPT-5 (w/ tools) & 51.5 & 26.2 & 73.0 & 54.8 & 60.9 & 46.2 \\
Gemini-2.5-Pro (w/ tools) & 38.5 & 25.3 & 59.4 & 47.3 & 51.3 & 38.4 \\
\midrule
OpenSearch-VL-8B~\citep{opensearchvl} & 11.2 & -- & 15.5 & 10.9 & 12.4 & -- \\
Qwen3-VL-8B~\citep{bai2025qwen3vltechnicalreport} & 13.6 & 5.5 & 9.6 & 15.5 & 13.5 & 10.9 \\
Qwen3-VL-32B~\citep{bai2025qwen3vltechnicalreport} & \textcolor{tablesecondgreen}{\underline{21.9}} & \textcolor{tablesecondgreen}{\underline{11.0}} & \textcolor{tablesecondgreen}{\underline{29.8}} & \textcolor{tablesecondgreen}{\underline{26.6}} & \textcolor{tablesecondgreen}{\underline{27.7}} & \textcolor{tablesecondgreen}{\underline{20.2}} \\
ModalSearcher-8B & \textcolor{tablebestblue}{\textbf{33.4}} & \textcolor{tablebestblue}{\textbf{21.3}} & \textcolor{tablebestblue}{\textbf{62.6}} & \textcolor{tablebestblue}{\textbf{37.4}} & \textcolor{tablebestblue}{\textbf{45.8}} & \textcolor{tablebestblue}{\textbf{33.5}} \\
\bottomrule
\end{tabularx}
\end{table}

\textbf{ModalSearcher transfers to independent visual-search benchmarks.} \revisiontext{Table~\ref{tab:interleave_benchs} combines independent visual-search benchmarks with ModalSearch-Bench to assess external transfer and performance under our task-construction paradigm.} \revisiontext{On the independently constructed VisBrowse and BrowseComp-V3 visual-search benchmarks, ModalSearcher-8B achieves} $33.4\%$ and $21.3\%$ accuracy, improving over Qwen3-VL-8B by $19.8$ and $15.8$ percentage points, respectively. It surpasses Qwen3-VL-32B by $11.5$ and $10.3$ points on these two benchmarks. \revisiontext{These consistent gains support transfer of active visual-search ability beyond our question-construction pipeline.} The tool ablation in Sec.~\ref{sec:ablation} shows that removing text-to-image search reduces accuracy on VisBrowse by $9.2$ points, supporting the role of actively acquired visual evidence on an independent benchmark. ModalSearch-Bench provides a controlled evaluation on held-out graph nodes, paths, and questions generated using the same construction procedure as the training data. \revisiontext{On ModalSearch-Bench, ModalSearcher-8B achieves $62.6\%$ on Input-only and $37.4\%$ on Search-required, yielding $45.8\%$ overall accuracy. The $21.9$-point gain over Qwen3-VL-8B on Search-required, together with improved visual-target alignment and coverage (Table~\ref{tab:visual-search-behavior}) and the visual-tool ablations (Figure~\ref{fig:ablation_framework}), supports our central motivation: training agents to actively acquire and use new visual information to advance reasoning.} \revisiontext{ModalSearch-Bench complements the external evaluations with a controlled assessment of active visual-information acquisition and use during reasoning. Its shared construction paradigm may contribute to the observed gains; the improvements on VisBrowse and BrowseComp-V3 provide evidence that the training benefit extends beyond our generated questions.}

\section{Ablation Studies}
\label{sec:ablation}

We conduct ablation studies from three perspectives: multimodal data generation, the agentic framework, and the training method.

\paragraph{\normalfont\textbf{Cross-modal data generation.}} The core design of \revisiontext{ModalSearch-Pipeline} is to sample reasoning paths containing different modality nodes from the interleaved multimodal knowledge graph. To validate the role of interleaved modality paths, we construct two data variants. For Input-only, we remove non-initial image nodes and retain visual evidence only at the beginning of the reasoning chain, then sample 7K examples for SFT with all other settings unchanged. \revisiontext{For Text-only, we remove all image nodes and separately sample 7K purely textual multi-hop search examples for SFT; these examples are excluded from the main SFT and RL data statistics.} For an equal-size comparison, the full modality configuration is also retrained on a randomly sampled 7K subset of the 13.5K pipeline-generated SFT examples. All three 7K settings use only generated data; the Full (16K) setting additionally includes the FVQA and LiveVQA training examples described in Sec.~\ref{sec:experiment-setup}. Results are reported in Table~\ref{tab:ablation_data}.

\begin{table}[!t]
\centering
\small
\setlength{\tabcolsep}{5pt}
\renewcommand{\arraystretch}{1.15}
\definecolor{tablebestblue}{RGB}{80,100,164}
\definecolor{tablesecondgreen}{RGB}{108,160,172}
\renewcommand{\textbf}[1]{{\bfseries\sffamily #1}}
\caption{Ablation of cross-modal data generation. All 7K variants use only pipeline-generated data; Full (16K) additionally includes FVQA and LiveVQA training examples.}
\label{tab:ablation_data}
\begin{tabular}{@{}lcccccc@{}}
\toprule
\textbf{Training data} & \textbf{MMSearch} & \textbf{BC-VL} & \textbf{VisBrowse} & \textbf{LiveVQA} & \textbf{ModalSearch} & \textbf{Avg.} \\
\midrule
Full (16K) & \textcolor{tablebestblue}{\textbf{64.3}} & \textcolor{tablebestblue}{\textbf{44.9}} & \textcolor{tablebestblue}{\textbf{28.9}} & \textcolor{tablesecondgreen}{\underline{68.3}} & \textcolor{tablebestblue}{\textbf{39.3}} & \textcolor{tablebestblue}{\textbf{49.1}} \\
Full (7K subset) & 63.0 & \textcolor{tablesecondgreen}{\underline{41.9}} & \textcolor{tablesecondgreen}{\underline{26.6}} & \textcolor{tablebestblue}{\textbf{69.1}} & \textcolor{tablesecondgreen}{\underline{37.3}} & \textcolor{tablesecondgreen}{\underline{47.6}} \\
Input-only (7K) & \textcolor{tablesecondgreen}{\underline{63.6}} & 39.8 & 21.3 & 67.4 & 33.0 & 45.0 \\
Text-only (7K) & 58.7 & 37.9 & 19.5 & 60.3 & 28.7 & 41.0 \\
\bottomrule
\end{tabular}
\end{table}

\begin{figure}[!t]
    \centering
    \edef\frameworkpanelheight{\the\dimexpr 0.239\linewidth\relax}
    \caption{Search-tool usage and agentic framework ablation. (a) Search-tool proportions, with raw invocation counts annotated. Red bars have a fixed width; each blue bar's width is scaled by the ModalSearcher-to-Qwen total-call ratio within that benchmark. Only \texttt{t2t\_search}, \texttt{t2i\_search}, and \texttt{i2i\_search} are counted; \texttt{read\_url} is excluded. (b) \revisiontext{Benchmark accuracy under progressively restricted tool settings, including no-tool answering;} each legend item pairs the Qwen3-VL-8B (red) and ModalSearcher-8B (blue) colors.}
    \label{fig:ablation_framework}
    \begin{minipage}[t]{0.28\linewidth}
        \vspace{0pt}
        \centering
        \includegraphics[height=\frameworkpanelheight]{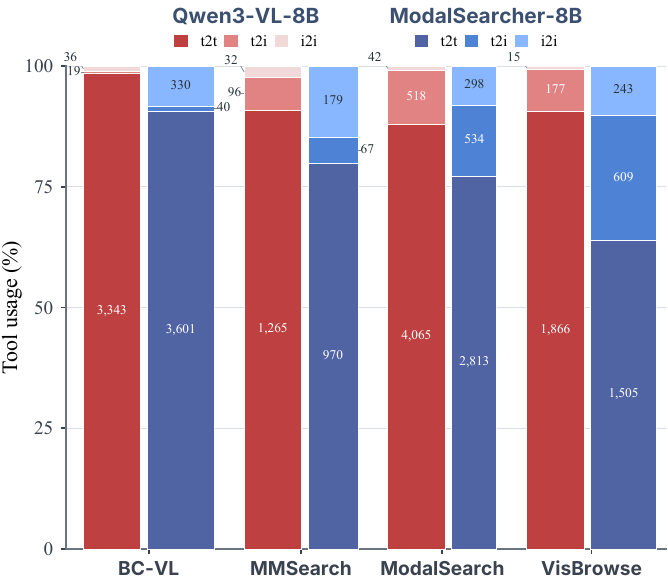}
        \par\smallskip\small (a) Search-tool usage
    \end{minipage}\hfill
    \begin{minipage}[t]{0.70\linewidth}
        \vspace{0pt}
        \centering
        \includegraphics[height=\frameworkpanelheight]{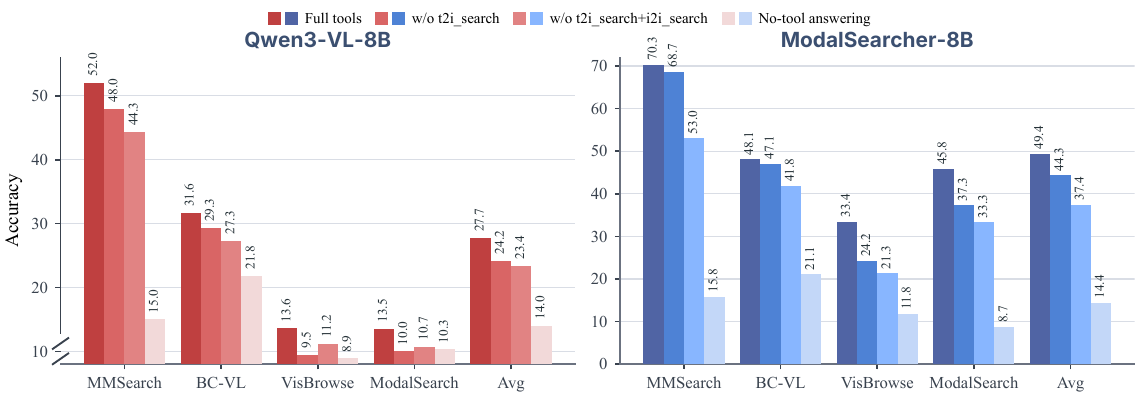}
        \par\smallskip\small (b) Tool ablation
    \end{minipage}
\end{figure}

Compared with the random Full (7K) subset, the Full (16K) training mixture improves ModalSearch (37.3$\rightarrow$39.3), BC-VL (41.9$\rightarrow$44.9), VisBrowse (26.6$\rightarrow$28.9), and Avg. (47.6$\rightarrow$49.1), while MMSearch and LiveVQA remain comparable. This comparison reflects both more generated examples and the addition of auxiliary training data, rather than isolating data scale alone. At the same 7K budget, Input-only retains similar performance on conventional multimodal QA but drops on ModalSearch (33.0), VisBrowse (21.3), and BC-VL (39.8), suggesting that an initial image alone is insufficient for these search tasks. Removing all image nodes in Text-only further degrades every benchmark, reducing ModalSearch to 28.7 and Avg. to 41.0. \revisiontext{These equal-size comparisons support the value of interleaved training data, with benefits on the external VisBrowse and BC-VL benchmarks as well as ModalSearch-Bench.}

\begin{table}[t]
    \begingroup
    \centering
    \small
    \setlength{\tabcolsep}{3pt}
    \renewcommand{\arraystretch}{1.15}
    \definecolor{tabledeltagreen}{HTML}{228B22}
    \renewcommand{\textbf}[1]{{\bfseries\sffamily #1}}
    \caption{Visual-search behavior before and after training on 200 search-required questions with 235 non-initial reference image nodes. All rates are percentages; $\Delta$ reports point differences computed before rounding. Alignment and coverage concern search targets rather than returned-image correctness.}
    \label{tab:visual-search-behavior}
    \begin{tabularx}{\linewidth}{@{}>{\raggedright\arraybackslash}p{0.25\linewidth}*{5}{>{\centering\arraybackslash}X}@{}}
        \toprule
        \textbf{Model} & \shortstack{\textbf{t2i}\\\textbf{invocation} $\uparrow$} & \shortstack{\textbf{Target}\\\textbf{alignment} $\uparrow$} & \shortstack{\textbf{New-target}\\\textbf{alignment} $\uparrow$} & \shortstack{\textbf{Node}\\\textbf{coverage} $\uparrow$} & \shortstack{\textbf{Repeated}\\\textbf{calls} $\downarrow$} \\
        \midrule
        Qwen3-VL-8B & 37.9 & 40.2 & 36.4 & 19.1 & 70.8 \\
        ModalSearcher-8B & \textbf{93.4} & \textbf{57.5} & \textbf{56.2} & \textbf{73.6} & \textbf{51.6} \\
        \midrule
        \cellcolor{gray!15}$\Delta$ vs Qwen3-VL-8B & \cellcolor{gray!15}\textcolor{tabledeltagreen}{+55.6} & \cellcolor{gray!15}\textcolor{tabledeltagreen}{+17.3} & \cellcolor{gray!15}\textcolor{tabledeltagreen}{+19.8} & \cellcolor{gray!15}\textcolor{tabledeltagreen}{+54.5} & \cellcolor{gray!15}\textcolor{tabledeltagreen}{-19.2} \\
        \bottomrule
    \end{tabularx}
\par\endgroup

    \vspace{8pt}
    \definecolor{tablebestblue}{RGB}{80,100,164}
    \definecolor{tablesecondgreen}{RGB}{108,160,172}
    \begin{minipage}[t]{0.47\textwidth}
        \vspace{0pt}
        \centering
        \captionsetup{type=table}
        \caption{Training ablation. \textcolor{tablebestblue}{\textbf{Bold}} and \textcolor{tablesecondgreen}{\underline{underline}} mark the best and second-best scores. MS/VB: ModalSearch/VisBrowse.}
        \label{tab:training_ablation}
        \smallskip
        \begingroup
\definecolor{tablebestblue}{RGB}{80,100,164}
\definecolor{tablesecondgreen}{RGB}{108,160,172}
\renewcommand{\textbf}[1]{{\bfseries\sffamily #1}}
\setlength{\tabcolsep}{2.6pt}
\renewcommand{\arraystretch}{1.392}
\begin{tabular}{@{}lccccc@{}}
\toprule
\textbf{Setting} & \textbf{MMSearch} & \textbf{BC-VL} & \textbf{MS} & \textbf{VB} & \textbf{Avg.} \\
\midrule
SFT+RL & \textcolor{tablebestblue}{\textbf{70.3}} & \textcolor{tablebestblue}{\textbf{48.1}} & \textcolor{tablebestblue}{\textbf{45.8}} & \textcolor{tablebestblue}{\textbf{33.4}} & \textcolor{tablebestblue}{\textbf{49.4}} \\
SFT-only & \textcolor{tablesecondgreen}{\underline{64.3}} & \textcolor{tablesecondgreen}{\underline{44.9}} & \textcolor{tablesecondgreen}{\underline{39.3}} & \textcolor{tablesecondgreen}{\underline{28.9}} & \textcolor{tablesecondgreen}{\underline{44.4}} \\
Base & 52.0 & 31.6 & 13.5 & 13.6 & 27.7 \\
\bottomrule
\end{tabular}
\endgroup

    \end{minipage}
    \hfill
    \begin{minipage}[t]{0.49\textwidth}
        \vspace{0pt}
        \centering
        \captionof{figure}{Accuracy and maximum-turn exceedance rate during RL training.}
        \label{fig:exp2_rl_dynamics}
        \smallskip
        \includegraphics[width=\linewidth]{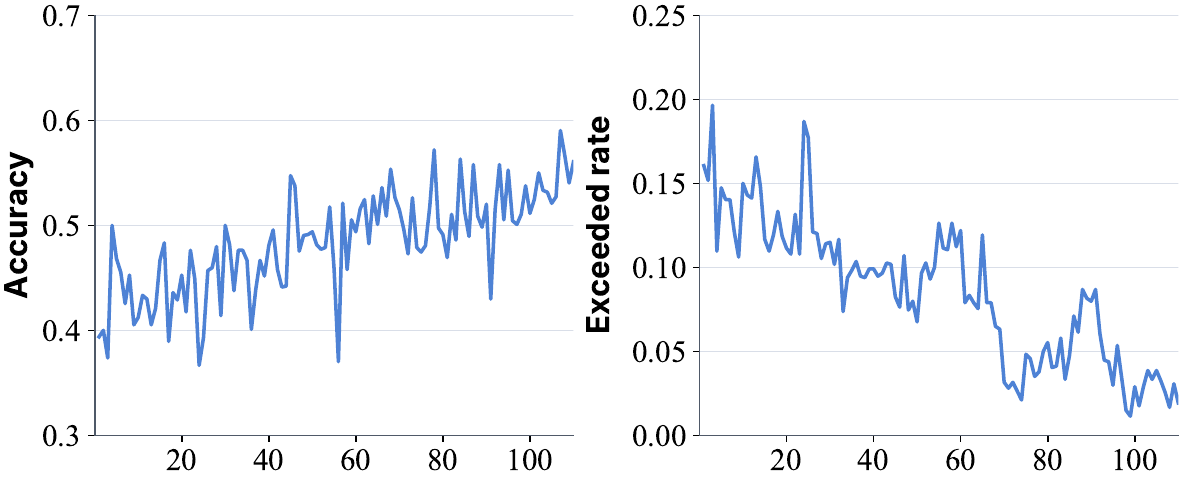}
    \end{minipage}
\end{table}

\paragraph{\normalfont\textbf{Agentic framework.}} \revisiontext{We test whether training improves the agent's use of actively retrieved visual information by combining tool-use statistics with tool-removal interventions. We compare Qwen3-VL-8B and ModalSearcher-8B under four settings: the full tool set, without \texttt{t2i\_search}, without both \texttt{t2i\_search} and \texttt{i2i\_search}, and no-tool answering. The two visual-tool ablations retain text search and URL reading. In the no-tool setting, search is unavailable, leaving no search-returned URLs for \texttt{read\_url}; the model answers from the provided input and its parametric knowledge.}

\revisiontext{Figure~\ref{fig:ablation_framework}(a) characterizes how training changes tool allocation. The base model predominantly uses \texttt{t2t\_search} across all four benchmarks. After training, the shares of both visual search tools increase on VisBrowse and ModalSearch. On MMSearch and BC-VL, reverse-image search also accounts for a larger share, while text-to-image search remains infrequent or does not increase in proportion. Total search calls decrease on MMSearch and ModalSearch but increase on BC-VL and VisBrowse, indicating that training changes tool allocation according to the task. Panel (b) tests whether access to these tools contributes to answer accuracy.}

\revisiontext{To examine whether visual searches target the reference evidence, Table~\ref{tab:visual-search-behavior} analyzes search-required questions in ModalSearch-Bench which contains 235 non-initial reference image nodes. Invocation rate measures the fraction of questions with at least one \texttt{t2i\_search} call. Target alignment is the fraction of calls whose search targets explicitly match a reference image node; new-target alignment restricts matches to non-initial nodes while retaining all calls as the denominator. Node coverage measures the fraction of distinct non-initial reference image nodes targeted by at least one call. Repeated-call rate measures the fraction of calls repeating the same search intent.}

\revisiontext{After training, visual-search invocation rises from $37.9\%$ to $93.4\%$, and non-initial target coverage increases from $19.1\%$ to $73.6\%$. Despite more calls overall, target alignment improves from $40.2\%$ to $57.5\%$, and the fraction of all calls aligned with non-initial visual targets rises from $36.4\%$ to $56.2\%$. The repeated-call proportion falls from $70.8\%$ to $51.6\%$. These results show that training increases targeted visual-search activity and reference-target coverage while reducing the proportion of repeated calls. The tool-removal experiment then tests whether visual-search access contributes to answering.}

\revisiontext{Removing \texttt{t2i\_search} reduces ModalSearcher's accuracy by $9.2$ and $8.5$ percentage points on VisBrowse and ModalSearch, respectively, compared with $4.1$ and $3.5$ points for the base model. Removing both visual search tools produces larger drops of $12.1$ and $12.5$ points for ModalSearcher, compared with $2.4$ and $2.8$ points for the base model. Thus, ModalSearcher derives a larger accuracy benefit from visual-search access on both benchmarks. With text search and URL reading still available, these accuracy drops provide direct evidence that visual-search access contributes to task performance. Together with Table~\ref{tab:visual-search-behavior}, they support improvements in both targeted visual-search behavior and the performance benefit of visual-search access. The corresponding effect on the independently constructed VisBrowse benchmark shows that this benefit is not confined to questions synthesized by ModalSearch-Pipeline.}

\revisiontext{The contribution also varies across benchmarks: removing \texttt{t2i\_search} costs ModalSearcher only $1.6$ and $1.0$ points on MMSearch and BC-VL, consistent with their predominantly textual search patterns in panel (a). Together, the tool-use and removal results support task-dependent use of visual search and its contribution to answering at the benchmark level; they do not assume that every question has an exclusively visual solution. Finally, no-tool answering reduces ModalSearcher's four-benchmark average from $49.4$ to $14.4$, close to the base model's $14.0$. The training benefit is therefore primarily reflected in search action and evidence integration.}

\paragraph{\normalfont\textbf{Training method.}} Table~\ref{tab:training_ablation} separates the contributions of SFT and RL. SFT alone substantially improves over the base Qwen3-VL-8B model, increasing MMSearch ($52.0 \rightarrow 64.3$), BrowseComp-VL ($31.6 \rightarrow 44.9$), and ModalSearch ($13.5 \rightarrow 39.3$). After initializing from the SFT model, RL further improves these benchmarks by $6.0$, $3.2$, and $6.5$ points, respectively. \revisiontext{These gains show that RL improves accuracy on both ModalSearch-Bench and external benchmarks, complementing the evidence for visual-tool utility in Figure~\ref{fig:ablation_framework}.} The fully trained model also reaches $33.4$ on VisBrowse, compared with $28.9$ after SFT and $13.6$ for the base model.

Figure~\ref{fig:exp2_rl_dynamics} shows answer accuracy on the training set and the fraction of rollouts truncated at the maximum interaction-turn limit during RL. We observe that, before RL, the model repeatedly issues semantically similar or even identical search queries when it fails to retrieve key clues, making little further reasoning progress. Early in RL training, the truncation rate peaks at nearly 20\%. As training progresses, answer accuracy generally increases while the truncation rate decreases. \revisiontext{These trends indicate that the agent becomes less prone to exhausting the interaction budget during training.} To further examine this behavioral change, we quantify repeated searching in Appendix~\ref{app:rl-details} and observe an overall decline in repeated-search frequency during RL. \revisiontext{These training-time observations are consistent with reduced redundant searching and complement the held-out accuracy gains in Table~\ref{tab:training_ablation}.}

\section{Conclusion}
\label{sec:conclusion}

We study autonomous cross-modal search for real-world information-seeking tasks that require interleaved textual and visual evidence. We introduce ModalSearch-Pipeline, a scalable data-generation framework that constructs an interleaved multimodal knowledge graph, samples paths with diverse depths and modality compositions, and \revisiontext{applies backward question rewriting to obscure entity names and visual targets that would expose reasoning shortcuts, while retaining necessary disambiguating clues and avoiding explicit modality-switch instructions.} Using the resulting data, we train the ModalSearcher Agent through SFT and RL and construct ModalSearch-Bench to evaluate cross-modal search across different modality arrangements and reasoning depths. ModalSearcher-8B improves average accuracy over Qwen3-VL-8B by $16.2$ points across five multimodal knowledge-intensive benchmarks and achieves substantial gains on independently constructed visual-search benchmarks. Behavioral analysis shows more targeted visual search, broader reference-target coverage, and a lower proportion of repeated calls after training. Together with data and tool ablations, these results support the value of the data and the contribution of active visual-search access to task performance.

\FloatBarrier

\clearpage
\bibliographystyle{plainnat}
\begingroup
\sloppy
\bibliography{main}
\endgroup

\clearpage
\appendix
\appendix
\section{Data Synthesis Details}
\label{app:data-synthesis}

This appendix supplements \revisiontext{ModalSearch-Pipeline} described in the main paper, including interleaved multimodal graph construction, path sampling, and question synthesis, and provides implementation details.

\begin{figure}[t]
    \centering
    \caption{Comparison of the data-construction strategies used by our pipeline and prior multimodal search settings. The comparison highlights differences in modality transitions, visual-evidence use, and the degree to which the search target is exposed in the question.}
    \label{fig:data_compare}
    \includegraphics[width=\linewidth]{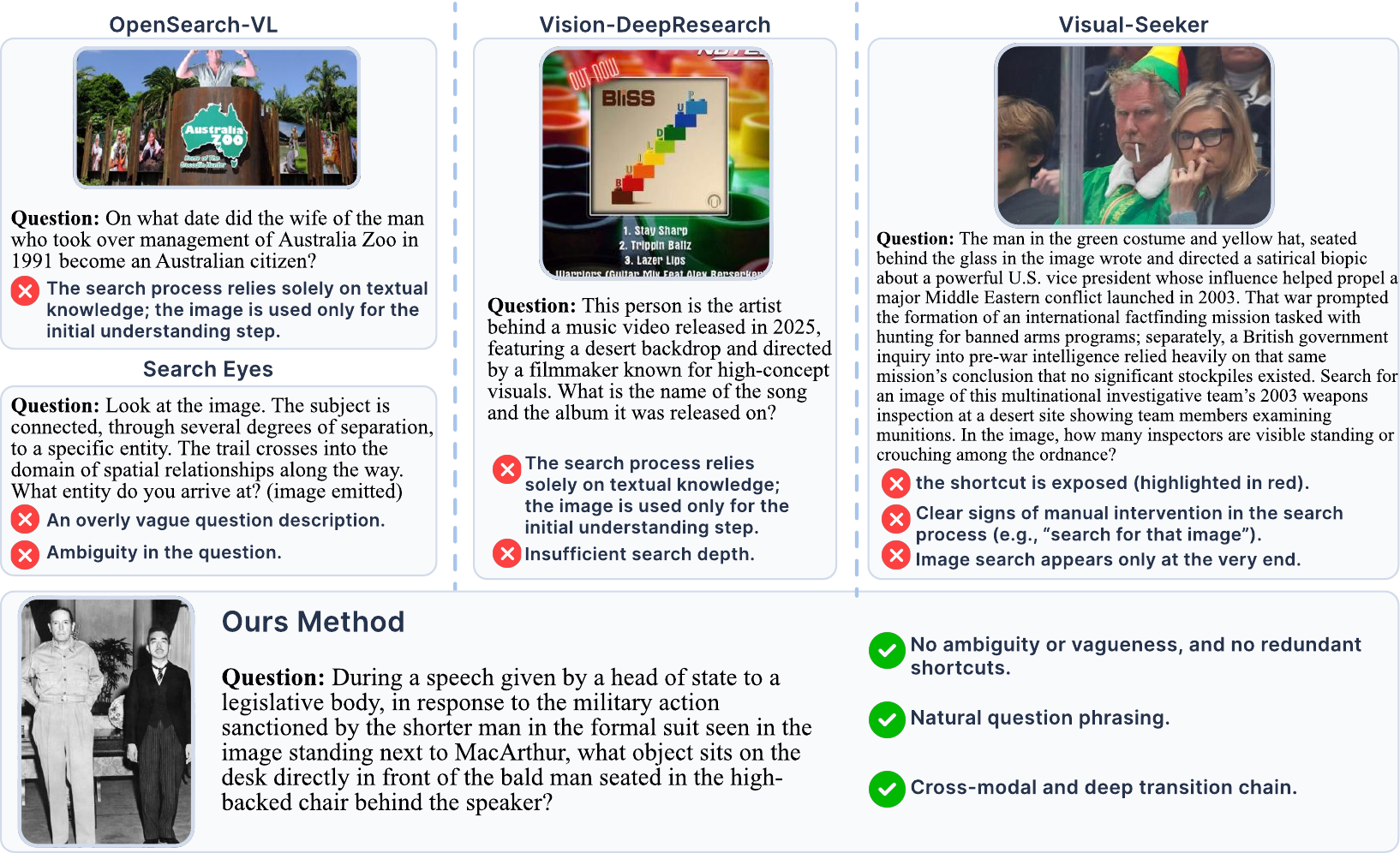}
\end{figure}

\subsection{Text-Node Expansion}
\label{app:text-expansion}

For any text node $v_{\mathrm{t}}=(a,w_a)$, where $a$ is the entity name and $w_a$ is the corresponding Wikipedia article, text-node expansion follows two complementary branches: relation-aware textual expansion extracts related textual entities, while retrievability-aware visual expansion proposes potential visual descriptions. Unlike directly using an existing Wikipedia knowledge graph, we randomly sample from hyperlinks and local textual contexts. This reduces dependence on frequent knowledge and common entity relations, increases path diversity, and makes it harder for an agent to solve the generated questions directly from parametric knowledge.

Specifically, we extract the local hyperlink structure from $w_a$ and apply rule-based filtering to remove non-entity pages, category pages, and clearly irrelevant links. We uniformly sample $K_{\mathrm{cand}}$ candidate entities from the remaining links, using $K_{\mathrm{cand}}=50$:
\begin{equation}
\widetilde{\mathcal{A}}(a)=\{\hat a_1,\ldots,\hat a_{K_{\mathrm{cand}}}\}.
\end{equation}
For each candidate $\hat a_j$, an MLLM (see Table~\ref{tab:data-generation-config}) uses the local context to identify the linked entity and generate the semantic relation between $a$ and $\hat a_j$. Because Wikipedia hyperlinks may be noisy---for example, they may point to non-entity pages, have weak semantic associations with the source entity, or be ambiguous---we further pass the candidate entities and their relation descriptions to an MLLM for quality filtering. According to relation relevance, relation clarity, and whether the target page denotes a concrete entity, we retain $K'$ high-quality entities from $\widetilde{\mathcal{A}}(a)$, where $K'\leq K_{\mathrm{cand}}$. For each retained entity, we construct $v_{\mathrm{t}}(\hat a_j)=(\hat a_j,w_{\hat a_j})$ and the edge $e_{\mathrm{t}\rightarrow\mathrm{t}}=(v_{\mathrm{t}}(a),r_{\mathrm{t}\rightarrow\mathrm{t}},v_{\mathrm{t}}(\hat a_j))$. Sampling only a limited number of candidates controls MLLM and graph-construction costs and prioritizes reasoning depth and richer paths over unconstrained graph width.

The text node simultaneously expands into neighboring image nodes. We divide $w_a$ into chunks $\mathcal{C}(w_a)=\{c_1,\ldots,c_m\}$. An MLLM examines each chunk and determines whether it contains visual content that can be matched to an Internet image, such as a specific scene, historical event, famous photograph, visual artwork, illustration, poster, cover, or film still. For qualifying chunks, the MLLM generates potential image descriptions or retrieval queries, forming a proposal set $\mathcal{P}(a)$. Since directly asking an MLLM whether an image is a valid downstream node is unstable, construction is divided into visual-description generation, referential-stability assessment, and search-result consistency verification.

Before image search, each $d\in\mathcal{P}(a)$ is screened by an MLLM for whether it can stably identify a particular type of visual content. For descriptions that pass this check, Web image search returns multiple candidate results, denoted by
\begin{equation}
\mathcal{R}(d)=\{(x_1,c_1),\ldots,(x_{K_{\mathrm{img}}},c_{K_{\mathrm{img}}})\},
\end{equation}
where $x_j$ is a candidate image and $c_j$ contains source URL, webpage title, image title, alt text, caption, and search summary. We first exclude query-mismatched or inaccessible images and form a valid candidate set for consistency assessment. Let $N(d)$ denote the number of images in this set, and let $M(d)$ denote the number judged by an MLLM (see Table~\ref{tab:data-generation-config}) to depict the same main visual content. We regard the description as stably retrievable when the consistency model returns a positive judgment and
\begin{equation}
N(d)\geq 2
\quad\text{and}\quad
M(d)\geq\max\!\left\{2,\left\lceil\frac{N(d)}{2}\right\rceil\right\}.
\end{equation}
The resulting image node is $v_{\mathrm{i}}=(x_i,d_i,c_i)$ with $d_i=d$; we then add $e_{\mathrm{t}\rightarrow\mathrm{i}}=(v_{\mathrm{t}}(a),r_{\mathrm{t}\rightarrow\mathrm{i}},v_{\mathrm{i}})$. Images that fail consistency verification are assigned to $V_{\mathrm{i}}^{\mathrm{amb}}$ and receive no incoming text-to-image edge, but may still serve as initial visual inputs.

Image referential stability means that a textual description can stably point to particular visual content rather than merely describing a broad theme. Under the same search setting, different searchers should obtain results concentrated on the same person, event, scene, or artwork. For event and scene images, the description should contain a time, location, or other clue that identifies a particular moment. For paintings, posters, covers, and film stills, it should identify a specific work or fixed visual object. Generic themes without an author or year, recurring events without distinctive temporal or spatial information, and descriptions that identify only a broad period or location, such as ``a CVPR conference in a certain year'' or ``a panorama of a bay,'' are treated as ambiguous. Because referential stability is partly subjective and difficult to formalize strictly, we annotate 116 image descriptions, comprising 63 stably retrievable positives and 53 retrieval-ambiguous negatives, and compare four MLLM judges against human annotations under the same rubric. Gemini-3.5-Flash achieves 88.79\% accuracy (Table~\ref{tab:uniqueness-judge}); the full comparison and judge-selection rationale are provided in Appendix~\ref{app:data-quality-analysis}. Ambiguous images do not receive text-to-image edges, but can be used as path-starting nodes because the initial image is provided to the agent.

\subsection{Image-Node Expansion}
\label{app:image-expansion}

For an image node $v_{\mathrm{i}}=(x_i,d_i,c_i)$, we perform verified visual entity linking to identify entities appearing in the image and expand them into downstream text nodes. Unlike standard image grounding, our graph-construction stage prioritizes entity recall so that potentially useful reasoning entities are not missed. We jointly input the original image $x_i$, its description $d_i$, and source-page metadata $c_i$ to an MLLM. The model uses visual content, the description, the source-page title, caption, alt text, search snippet, and related context to identify possible entities and generate uniquely referring descriptions. The preliminary entity set is
\begin{equation}
\mathcal{A}(x_i)=\{a'_1,\ldots,a'_n\}.
\end{equation}

For each candidate $a'$, we first perform Wikipedia entity linking. We construct entity-search queries from the proposed name, entity type, visual evidence, image description, and source-page title, and call the MediaWiki API to retrieve a candidate set $\mathcal{U}(a')$. For each candidate article, we obtain its title, summary, canonical URL, and available Wikidata QID. Candidates are initially ranked by string and lexical overlap between the entity description and candidate title or summary. The candidate list and its context are then passed to an MLLM for disambiguation. If candidates remain ambiguous or the evidence is insufficient, we discard the match rather than forcing a link.

For a selected Wikipedia article, we reuse the corresponding text node if its canonical URL already exists; otherwise, we add the article to the text-node expansion queue. We then jointly input the original image, the article content, and reference images to a verification model. Through cross-image comparison, the model determines whether the entity in the original image corresponds to the selected article. Let $\mathcal{A}^{\star}(x_i)\subseteq\mathcal{A}(x_i)$ be the verified entity set. For each $a'\in\mathcal{A}^{\star}(x_i)$, we construct $v_{\mathrm{t}}(a')=(a',w_{a'})$ and $e_{\mathrm{i}\rightarrow\mathrm{t}}=(v_{\mathrm{i}},r_{\mathrm{i}\rightarrow\mathrm{t}},v_{\mathrm{t}}(a'))$, where $r_{\mathrm{i}\rightarrow\mathrm{t}}$ describes the visual or semantic relation between $a'$ and the image, such as its position, event participation, scene membership, or relation to the recorded artwork. This process forms the typical cross-modal path $v_{\mathrm{t}}\rightarrow v_{\mathrm{i}}\rightarrow v'_{\mathrm{t}}$. Entity recognition, entity linking, and visual verification solve different problems. Recognition aims to discover candidate entities with high recall; linking maps a natural-language description to a concrete Wikipedia article; and verification checks whether that article actually corresponds to the entity in the image. Only candidates that pass both entity linking and cross-image verification are included in the final image-to-text relation set, reducing the effect of erroneous matches on path sampling and question generation.

\subsection{Graph Assembly and Node Reuse}

We randomly select multiple initial nodes and begin constructing the interleaved multimodal graph. Before scheduling a node-expansion task, a hash table checks whether the node already exists in the current graph. For a new node, we add its expansion task to a waiting queue and create a deferred edge-completion task. Once the node is actually expanded, the deferred task adds the directed edge between the new node and its parent. This procedure reuses existing nodes and prevents duplicate expansions.

\subsection{Graph Statistics}
\label{app:graph-statistics}

\begin{figure}[t]
    \centering
    \caption{Overview of the pipeline-generated training data. \revisiontext{Left: counts and proportions of Input-only examples, whose reference-path visual evidence is provided initially, and Search-required examples, whose reference solutions call for actively acquiring new visual information during reasoning.} Right: a word cloud of 70 randomly sampled entities, illustrating the breadth of covered domains. \revisiontext{Auxiliary FVQA and LiveVQA training examples and the separately sampled Text-only ablation data are excluded.}}
    \label{fig:training-data-overview}
    \includegraphics[width=0.32\linewidth]{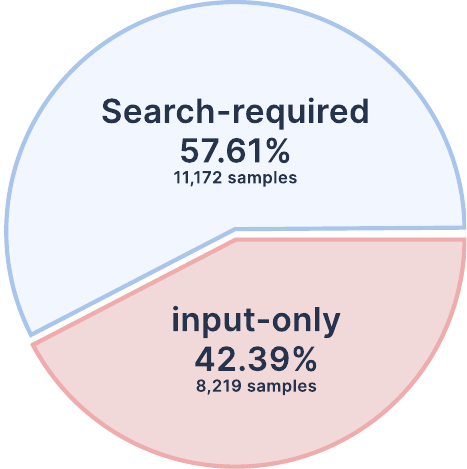}\hfill
    \includegraphics[width=0.65\linewidth]{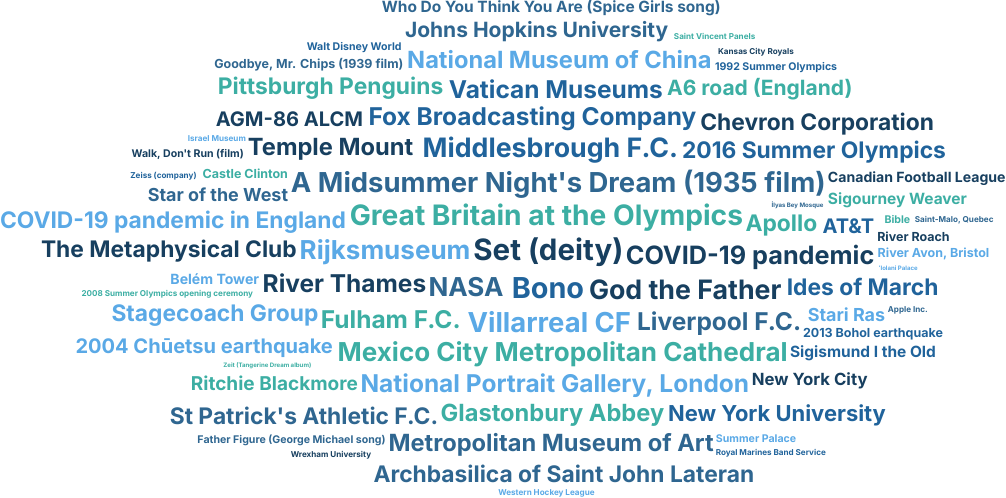}
\end{figure}

Table~\ref{tab:graph-statistics} summarizes the source graphs used for training-data and benchmark construction. The training graph contains 70,048 nodes and 144,511 valid directed edges, while the independently constructed benchmark graph contains 20,035 nodes and 40,042 valid directed edges. The image-category counts use the merged labels from the graph records, with previously unclassified images included in the non-unique category. The path-length distribution of the generated training data is reported in Table~\ref{tab:path-length-distribution}, and the modality-path overview is shown in Figure~\ref{fig:training-data-overview}.

\begin{table}[!t]
\centering
\small
\setlength{\tabcolsep}{10pt}
\renewcommand{\arraystretch}{1.12}
\renewcommand{\textbf}[1]{{\bfseries\sffamily #1}}
\caption{Statistics of the source graphs for training-data and benchmark construction. Unique images combine the original \texttt{unique} and \texttt{semi-unique} categories; non-unique images combine \texttt{no-unique}, \texttt{wiki\_inline}, and previously unclassified images.}
\label{tab:graph-statistics}
\begin{tabular}{@{}lrr@{}}
\toprule
\textbf{Metric} & \textbf{Training graph} & \textbf{Benchmark graph} \\
\midrule
Text nodes & 35{,}000 & 10{,}000 \\
Image nodes & 35{,}048 & 10{,}035 \\
Total nodes & 70{,}048 & 20{,}035 \\
\midrule
Text-to-text edges & 53{,}811 & 20{,}386 \\
Text-to-image edges & 19{,}546 & 5{,}915 \\
Image-to-text edges & 71{,}154 & 13{,}741 \\
Total valid edges & 144{,}511 & 40{,}042 \\
\midrule
Unique images & 16{,}261 & 5{,}814 \\
Non-unique images & 18{,}787 & 4{,}221 \\
\bottomrule
\end{tabular}
\end{table}

\subsection{Path Sampling}
\label{app:path-sampling}

On $G=(V,E)$, path sampling starts from a random node $v_0$ and samples a target reasoning depth $L\in[K_{\min},K_{\max}]$, with $[K_{\min},K_{\max}]=[1,4]$. At step $t$, the sampler obtains the outgoing edges of $v_t$ and constructs
\begin{equation}
\mathcal{C}_t=\{e=(v_t,r,v')\mid v'\notin H_t,\ e\notin U_t,\ \phi(e,H_t)=1\},
\end{equation}
where $H_t=(v_0,\ldots,v_t)$ is the path history, $U_t$ is the set of already used edges, and $\phi$ is the candidate-validity function. It imposes graph-structural and modality constraints, avoids repeated nodes and edges, filters edges leading to generalized category pages, and excludes structural shortcuts to historical nodes. Qwen3-VL-32B serves as the path-validity checking model and checks whether the target name $v'$ has already been exposed in a relation description along $H_t$; if so, $\phi(e,H_t)=0$.

The sampler computes a base weight from the edge's historical usage and image spacing. For edge $e$, $w(e\mid H_t)=\gamma_{\mathrm{img}}(e,H_t)/(1+\alpha c(e))$, where $c(e)$ is the number of previously accepted paths using $e$ and $\alpha$ controls the repetition penalty. Let $j^{\star}=\max\{j\mid v_j\in V_{\mathrm{i}}\}$ be the most recent image position in the history and $d_t=t-j^{\star}$. To avoid densely clustered image nodes, we use
\begin{equation}
\gamma_{\mathrm{img}}(e,H_t)=
\begin{cases}
\dfrac{d_t}{d_t+1}, & v'\in V_{\mathrm{i}},\ \text{an image occurs in }H_t,\ \text{and }d_t>0,\\
1, & \text{otherwise}.
\end{cases}
\end{equation}
The next edge is sampled with $p(e\mid H_t)=w(e\mid H_t)/\sum_{e'\in\mathcal{C}_t}w(e'\mid H_t)$ rather than by always selecting the highest-weight edge, preserving trajectory diversity. After a path is completed, edge-use counts are updated and duplicate paths are removed using their complete node sequences.

The goal of the pipeline is to train the agent to autonomously select the modality of the clue it needs, rather than to follow explicit instructions to search for an image or text fact. Before question synthesis, an MLLM therefore scans intermediate image nodes and combines the two natural-language edge descriptions in each text--image--text subpath into a composed relation description. This description connects the two text nodes for question synthesis and refers to the relevant event, scene, or artwork without explicitly mentioning the image. The original image-mediated subpath is retained as the supporting evidence. This rewriting is intended to make the visual evidence dependency implicit and encourage the agent to determine which modality to search and what content to retrieve; it does not by itself rule out alternative textual solution paths.

\subsection{Question Synthesis}
\label{app:question-writing}

Question synthesis uses backward question rewriting, a recursive process that proceeds from the end of a reasoning path toward its start. Given $\pi=(v_0,v_1,\ldots,v_L)$, we first generate a terminal question according to the modality of $v_L$. If $v_L=v_{\mathrm{t}}(a)=(a,w_a)$, an MLLM reads $w_a$ and generates $k_2$ candidate questions. Each candidate contains a question $q$, answer $y$, and supporting evidence $\xi$; the question must not directly reveal the target entity, an explicit alias, or another identifying description. If $v_L=v_{\mathrm{i}}=(x_i,d_i,c_i)$, the MLLM generates $k_2$ questions about observable visual attributes or relations while minimizing dependence on source-page metadata or external semantic knowledge.

We apply modality-specific question evaluation, with answer-based verification when a verification model $M_{\mathrm{v}}$ is configured; we use Qwen3-VL-32B as $M_{\mathrm{v}}$. For a text terminal node, $M_{\mathrm{v}}$ answers each candidate without access to the Wikipedia article or other external evidence. Its responses are judged against the proposed gold answers, and candidates answered correctly in this closed-book test are removed. A subsequent evaluator selects a natural, unambiguous question supported by its article-derived facts. This branch does not independently test whether $M_{\mathrm{v}}$ can answer the question when given the article.

For an image terminal node, answer-based verification performs $T=3$ trials per condition, with and without the reference image. Let $C_{\mathrm{with}}(q)$ and $C_{\mathrm{without}}(q)$ denote the numbers of correct responses in the two conditions. Candidates pass this verification when
\begin{equation}
C_{\mathrm{with}}(q)=T
\quad\text{and}\quad
C_{\mathrm{without}}(q)<T,\qquad T=3.
\end{equation}
A subsequent evaluator checks referent uniqueness, answer stability, and visual correctness. Among candidates that pass both verification and evaluation, we prefer those with fewer correct responses without the image. This favors questions that are reliably answerable with visual evidence and less consistently answerable from parametric knowledge alone. The evaluator may reject all candidates; otherwise, its selected candidate becomes the initial terminal question $q_L$.

We then recursively rewrite the question backward. For each edge $e_n=(v_{n-1},r_n,v_n)$, the LLM removes redundant or path-irrelevant content from $r_n$, obtaining $\bar r_n$, and replaces the explicit reference to $v_n$ in $q_n$ with a relational description defined by $v_{n-1}$ and $\bar r_n$. Formally, $q_{n-1}=\mathcal{R}(q_n;v_{n-1},\bar r_n,v_n)$, where $\mathcal{R}$ is the relation-based iterative rewriting operation. \revisiontext{Each rewrite obscures explicit intermediate-node references that would enable reasoning shortcuts while preserving names and clues needed for solvability and answer uniqueness.}

After rewriting for $n=L,L-1,\ldots,1$, the question $q_0$ explicitly contains only the starting node and necessary reasoning clues. For a textual start node $v_0=v_{\mathrm{t}}(a)=(a,w_a)$, we input $w_a$ to an MLLM (see Table~\ref{tab:data-generation-config}) and ask it to generate an indirect description $\tilde a$ that does not reveal the entity name; replacing $a$ with $\tilde a$ yields $\tilde q_0$. For an image start node $v_0=v_{\mathrm{i}}=(x_0,d_0,c_0)$, we replace the image description with ``this image'' and provide $x_0$ as the visual input. \revisiontext{The final question $\tilde q_0$ hides entity names and image-search targets when their disclosure would enable reasoning shortcuts, while retaining names and clues necessary for solvability and answer uniqueness. It avoids explicit modality-switch instructions and is designed to encourage the agent to identify its knowledge gaps, plan searches, and integrate evidence across multiple hops.} Figure~\ref{fig:data_compare} provides a qualitative comparison with existing question-construction strategies. The generated data (i) embeds image nodes in multi-hop reasoning paths, (ii) balances shortcuts caused by redundant information against ambiguity caused by insufficient information, and \revisiontext{(iii) avoids directly exposing shortcut-enabling search targets or prescribing a fixed reasoning trajectory.}

\section{ModalSearch-Bench Details}
\label{app:benchmark-details}

\begin{table}[!t]
\centering
\small
\setlength{\tabcolsep}{6pt}
\renewcommand{\arraystretch}{1.12}
\renewcommand{\textbf}[1]{{\bfseries\sffamily #1}}
\caption{Models and parameters used in \revisiontext{ModalSearch-Pipeline}. Answer verification is distinct from benchmark answer judging. Visual verification uses three trials per condition, with and without the reference image.}
\label{tab:data-generation-config}
\begin{tabularx}{\linewidth}{@{}>{\raggedright\arraybackslash}p{0.17\linewidth} >{\raggedright\arraybackslash}X >{\raggedright\arraybackslash}p{0.24\linewidth}@{}}
\toprule
\textbf{Stage} & \textbf{Configuration} & \textbf{Value / Setting} \\
\midrule
Graph construction & Relation-aware textual expansion model & Qwen3-VL-32B \\
& Visual expansion, entity recognition, and linking model & Gemini-2.5-Pro \\
& Image referential-stability assessment model & Gemini-3.5-Flash \\
& Textual neighbor candidate budget $K_{\mathrm{cand}}$ & 50 \\
& Maximum visual proposals per text node & 5 \\
& Retrieved images per query $K_{\mathrm{img}}$ & 3 \\
& Maximum grounded entities per image & 5 \\
\midrule
Path sampling & Path-validity checking model & Qwen3-VL-32B \\
& Path-length range $[K_{\min},K_{\max}]$ & $[1,4]$ \\
& Edge-reuse penalty coefficient $\alpha$ & 1.0 \\
\midrule
Question synthesis & Question synthesis model & Gemini-2.5-Pro \\
& Answer-verification model $M_{\mathrm{v}}$ & Qwen3-VL-32B \\
& Terminal-question candidates per node $k_2$ & 5 \\
& Visual answer-verification trials $T$ & 3 per condition \\
\bottomrule
\end{tabularx}
\end{table}

ModalSearch-Bench is designed to evaluate multimodal deep-search agents on diverse autonomous cross-modal search problems. Unlike benchmarks that organize questions primarily into coarse difficulty levels, \revisiontext{it categorizes questions by whether their reference solutions call for acquiring new visual information beyond the initial input.} \revisiontext{In Input-only questions, all visual evidence in the reference solution is provided with the question.} Search-required questions \revisiontext{call for the agent to actively search for new visual information during reasoning and use it to advance the solution, with or without an initial image.} \revisiontext{These labels characterize the reference evidence paths rather than exclude alternative textual solutions or restrict the tools an agent may invoke.} Within each category, questions are grouped into 2-hop, 3-hop, and 4-hop subsets to evaluate different search depths.

Following the construction procedure in Sec.~\ref{sec:method}, we first use a different set of initial seed nodes to build a new interleaved multimodal knowledge graph with 20,035 nodes. Its seed nodes are entirely disjoint from those of the graph used to construct the training data, and some belong to domains not covered by the training graph. Using the same path-sampling and question-synthesis procedure, we randomly sample 900 candidate questions spanning different visual-evidence arrangements and the three reasoning depths. Human experts then screen these candidates and retain 300 questions. For every retained example, they consult Wikipedia to verify the answer, correctness, and solvability of the question and to remove ambiguous cases. The benchmark contains 100 Input-only and 200 Search-required questions, with a broadly balanced distribution across 2-hop, 3-hop, and 4-hop questions. Overall accuracy is computed over all 300 questions rather than by equally averaging the two category accuracies. All benchmark graph nodes, paths, and questions are held out from training. ModalSearch-Bench therefore provides a controlled evaluation of a multimodal deep-search agent's ability to plan multi-hop searches, select and switch modalities, and integrate cross-modal evidence.

\section{Implementation Details}
\label{app:implementation-details}

We use LlamaFactory for supervised fine-tuning (SFT), and the rLLM and verl frameworks for reinforcement learning (RL). During SFT, trajectories are collected with GPT-5.4 given a reasoning path and the final answer. The model autonomously invokes the available tools and produces trajectories in the ReAct format. We describe the training data and implementation settings below.

\subsection{Training Data Details}
\label{app:training-data-details}

For the data-construction pipeline described in Sec.~\ref{sec:method} and Appendix~\ref{app:data-synthesis}, the main parameters and configuration are summarized in Table~\ref{tab:data-generation-config}. Using \revisiontext{ModalSearch-Pipeline} and the interleaved multimodal knowledge graph described in Appendix~\ref{app:graph-statistics}, we sample 13,520 examples for SFT and 5,871 additional examples for RL (reported as 5.8K in the main text). The full SFT set contains approximately 16K examples after adding 1.3K FVQA and 1.2K LiveVQA training examples, as described in Sec.~\ref{sec:experiment-setup}. Table~\ref{tab:path-length-distribution} reports hop counts for the pipeline-generated examples only, excluding these auxiliary datasets. \revisiontext{The dataset spans one to four graph hops to cover different search depths and modality arrangements; graph-hop counts are distinct from agent interaction turns.} We group multimodal examples into Input-only and Search-required, using the same evidence-based definitions as ModalSearch-Bench in Appendix~\ref{app:benchmark-details}. Input-only examples contain visual evidence only at the start; \revisiontext{Search-required examples call for actively acquiring new visual information during reasoning and using it to advance the solution.} The generated-data overview comprises 8,219 Input-only examples (42.39\%) and 11,172 Search-required examples (57.61\%). This grouping distinguishes input-image understanding from active acquisition of new visual evidence. \revisiontext{The 19,391 generated examples used for the main SFT and RL experiments are all multimodal. Purely textual examples are sampled separately using the same pipeline for the Text-only ablation and are excluded from these counts.} The contribution of these data settings is evaluated in Sec.~\ref{sec:ablation}.

\begin{table}[!t]
\centering
\small
\setlength{\tabcolsep}{10pt}
\renewcommand{\arraystretch}{1.12}
\renewcommand{\textbf}[1]{{\bfseries\sffamily #1}}
\caption{Reasoning-path length distribution of pipeline-generated training examples. \revisiontext{The separately sampled Text-only ablation data and the FVQA and LiveVQA examples added to SFT are excluded.} Share is computed over the combined SFT and RL total.}
\label{tab:path-length-distribution}
\begin{tabular}{@{}lrrrr@{}}
\toprule
\textbf{Hops} & \textbf{SFT} & \textbf{RL} & \textbf{Total} & \textbf{Share} \\
\midrule
1 & 4{,}493 & 1{,}412 & 5{,}905 & 30.45\% \\
2 & 4{,}721 & 2{,}001 & 6{,}722 & 34.67\% \\
3 & 3{,}545 & 1{,}760 & 5{,}305 & 27.36\% \\
4 & 761 & 698 & 1{,}459 & 7.52\% \\
\midrule
Total & 13{,}520 & 5{,}871 & 19{,}391 & 100\% \\
\bottomrule
\end{tabular}
\end{table}

\FloatBarrier
\subsection{SFT Details}
\label{app:sft-details}

The SFT hyperparameters in the provided training configuration are summarized in Table~\ref{tab:sft-config}. The YAML specifies full-parameter fine-tuning of Qwen3-VL-8B with DeepSpeed ZeRO-3 and 16 Ray workers. With a per-device batch size of one and 16 accumulation steps, the effective batch size is 256; the cosine scheduler uses 0.5 cycles and decays to zero over four epochs.

\subsection{RL Details}
\label{app:rl-details}

Starting from the SFT checkpoint, we train on 5.8K additional examples with RLOO for four epochs. The run uses an outcome-only reward with a $0.9$ answer-accuracy weight and a $0.1$ format-validity weight; partial-credit and advantage-shaping mechanisms are disabled. The key optimization and rollout settings are summarized in Table~\ref{tab:rl-config}.

\begin{table}[!t]
\centering
\renewcommand{\arraystretch}{1.06}
\begin{minipage}[t]{0.485\textwidth}
\vspace{0pt}
\footnotesize
\centering
\captionof{table}{Selected SFT training configuration.}
\label{tab:sft-config}
\begin{tabular}{@{}p{0.55\linewidth} p{0.35\linewidth}@{}}
\toprule
\textbf{Hyperparameter} & \textbf{Value / Setting} \\
\midrule
\multicolumn{2}{@{}l}{\textbf{Model}} \\
Base model & Qwen3-VL-8B \\
Image max pixels & $262{,}144$ ($\approx 512 \times 512$) \\
Flash Attention & \texttt{fa2} \\
\midrule
\multicolumn{2}{@{}l}{\textbf{Method}} \\
Stage & SFT \\
Do train & True \\
Finetuning type & Full \\
Vision tower frozen & False \\
MM projector frozen & False \\
DeepSpeed & ZeRO-3 \\
Mixed precision & bfloat16 \\
\midrule
\multicolumn{2}{@{}l}{\textbf{Training}} \\
Batch size per device & 1 \\
Effective batch size & 256 \\
Learning rate & $2.0 \times 10^{-5}$ \\
Epochs & 4 \\
LR scheduler & Cosine \\
Scheduler cycles & $0.5$ \\
Warmup ratio & 0.1 \\
\bottomrule
\end{tabular}
\end{minipage}
\hfill
\begin{minipage}[t]{0.485\textwidth}
\vspace{0pt}
\footnotesize
\centering
\captionof{table}{Selected RL training configuration.}
\label{tab:rl-config}
\begin{tabular}{@{}p{0.55\linewidth} p{0.35\linewidth}@{}}
\toprule
\textbf{Hyperparameter} & \textbf{Value / Setting} \\
\midrule
\multicolumn{2}{@{}l}{\textbf{Model}} \\
Starting checkpoint & SFTed Qwen3-VL-8B \\
\midrule
\multicolumn{2}{@{}l}{\textbf{Data}} \\
Training prompt batch size & 96 (mini-batch: 96) \\
Total training examples & 5.8K \\
\midrule
\multicolumn{2}{@{}l}{\textbf{Algorithm / Reward}} \\
Advantage estimator & RLOO \\
Accuracy coefficient & 0.9 \\
Format coefficient & 0.1 \\
\midrule
\multicolumn{2}{@{}l}{\textbf{Training}} \\
Total epochs & 4 \\
Loss aggregation & Token mean \\
\midrule
\multicolumn{2}{@{}l}{\textbf{Rollout}} \\
Rollout engine & vLLM \\
GPU memory utilization & 0.80 \\
Tensor parallel size & 1 \\
Max batched tokens & 8{,}192 \\
Parallel tasks / tools & 768 / 768 \\
Responses per prompt & 6 \\
\bottomrule
\end{tabular}
\end{minipage}
\end{table}

\FloatBarrier
To improve tool-call stability during RL and reduce tool-use cost, we cache tool results within each rollout group. Specifically, when a search tool is reused within a group for the same query, we directly return the cached search result instead of invoking the tool again. In the actual training runs, the cache hit rate reaches 18--22\%.

Because the generated rollouts are long and search or page-fetching tools can become unstable under high concurrency due to network or service failures, we introduce a set of fallback mechanisms. If an individual format check or tool call is temporarily unavailable, the tool returns a textual notice indicating the corresponding failure type. If errors occur for three consecutive rounds during reasoning, we attribute the failure. When the error is caused by the agent action, such as an invalid model format or malformed tool parameters, we terminate the rollout and set the sample reward to $0.1$. When the error is caused by tool instability, we also terminate the rollout but set the reward to $0$. These erroneous samples still participate in the within-group advantage computation.

Before RL, the model may repeatedly issue semantically similar or identical search queries when it fails to retrieve key clues, making little further reasoning progress until the rollout is truncated by the maximum-turn limit. To quantify this behavior, we track two statistics throughout RL training: (i) the fraction of rollouts terminated by the maximum-turn cap, and (ii) the average number of repeated-search events. For a rollout $\tau$, with $L(\tau)$ interaction turns and maximum allowed length $L_{\max}$, the truncation indicator and the training-step truncation rate are
\begin{equation}
    \mathbb{I}_{\mathrm{cap}}(\tau)=\mathbb{I}[L(\tau)\geq L_{\max}],\qquad
    C_t=\frac{1}{N_t}\sum_{n=1}^{N_t}\mathbb{I}_{\mathrm{cap}}(\tau_n),
\end{equation}
where $N_t$ rollouts are collected at training step $t$.

\begin{figure}[t]
    \centering
    \caption{RL training dynamics over 110 training steps: (a) answer accuracy on the training set, (b) the fraction of rollouts truncated at the maximum interaction-turn limit, and (c) the average number of repeated-search events per rollout. All panels show the same training run (Experiment 2). A repeated search is counted when two queries have cosine similarity above $0.9$ and at least three of their five returned results overlap.}
    \label{fig:rl_redundancy}
    \includegraphics[width=\linewidth]{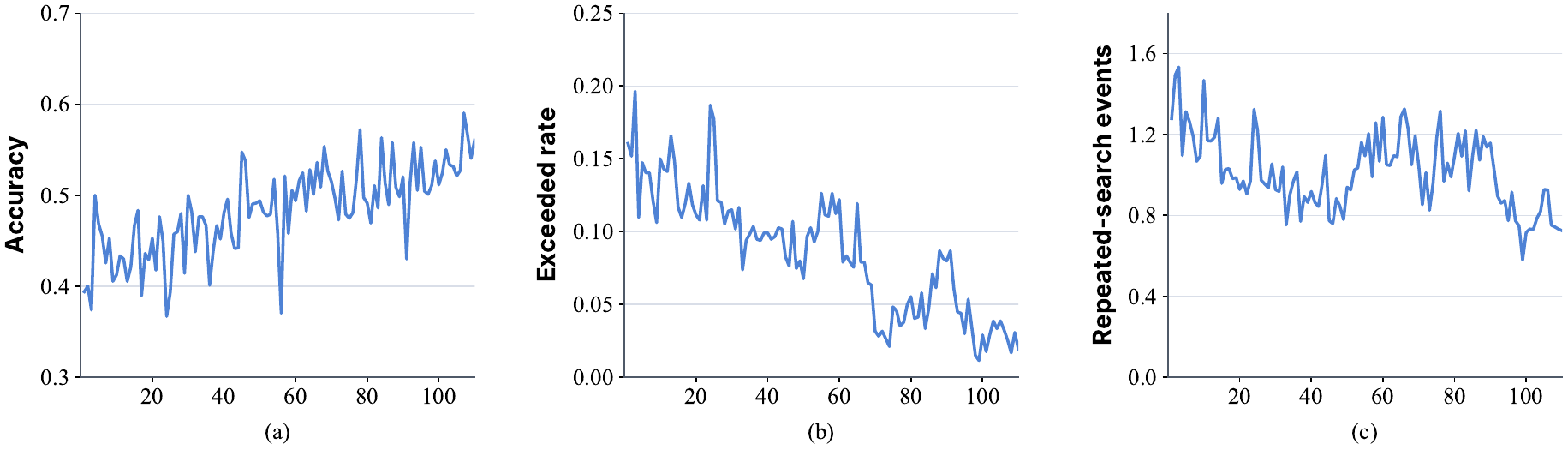}
\end{figure}

For the repeated-search statistic, consider two search operations $s_1=(q_1,S_1)$ and $s_2=(q_2,S_2)$, where $q_i$ is the issued query and $S_i$ contains the five returned search results. We encode each query with Qwen3-Embedding-4B, denoting its feature vector by $z_i=f_{\mathrm{emb}}(q_i)$, and compute query similarity as
\begin{equation}
    \operatorname{sim}_{q}(s_1,s_2)=\frac{z_1^{\top}z_2}{\lVert z_1\rVert_2\lVert z_2\rVert_2}.
\end{equation}
We measure result overlap by the fraction of identical items among the five results,
\begin{equation}
    \operatorname{ov}(s_1,s_2)=\frac{\lvert S_1\cap S_2\rvert}{5}.
\end{equation}
We classify the two operations as a repeated search when both the query similarity and the result overlap exceed their respective thresholds:
\begin{equation}
    \mathbb{I}_{\mathrm{rep}}(s_1,s_2)=\mathbb{I}\!\left[\operatorname{sim}_{q}(s_1,s_2)>0.9\;\land\;\operatorname{ov}(s_1,s_2)>0.4\right].
\end{equation}
Since each result set contains five items, the latter condition corresponds to at least three identical results (i.e., more than two matches). For rollout $\tau$, the repeated-search count is $R(\tau)=\sum_{(s_i,s_j)\in\mathcal{P}(\tau)}\mathbb{I}_{\mathrm{rep}}(s_i,s_j)$, where $\mathcal{P}(\tau)$ denotes the unordered pairs of search operations in that rollout. We report the training-step average $\bar R_t=\frac{1}{N_t}\sum_{n=1}^{N_t}R(\tau_n)$ over the $N_t$ rollouts collected at step $t$. The evolution of this statistic is shown in Figure~\ref{fig:rl_redundancy}.

Figure~\ref{fig:rl_redundancy} shows training answer accuracy, the maximum-turn truncation rate, and the average number of repeated-search events over 110 RL training steps. Despite fluctuations, answer accuracy shows an overall increase, while both the truncation rate and repeated-search count show an overall decline. Together, these trends support the conclusion that RL improves multi-round search decisions, reduces redundant searching, and makes the agent less prone to unproductive long rollouts.

\FloatBarrier
\subsection{Inference Configuration}
\label{app:inference-configuration}

Unless otherwise specified, all models are evaluated using the sampling parameters summarized in Table~\ref{tab:sampling-params}. We use GPT-4o to judge answer correctness with the evaluation prompt in Figure~\ref{fig:prompt_evaluation}. Of the evaluated model's output, only the final summarized answer is passed to the judge; the full reasoning trajectory is excluded.

\begin{table}[!ht]
\centering
\small
\setlength{\tabcolsep}{8pt}
\renewcommand{\arraystretch}{1.1}
\caption{Sampling parameters used for inference.}
\label{tab:sampling-params}
\begin{tabular}{@{}ll@{\hspace{8pt}\vrule width 0.4pt\hspace{8pt}}ll@{}}
\toprule
\textbf{Parameter} & \textbf{Value} & \textbf{Parameter} & \textbf{Value} \\
\midrule
\texttt{greedy}              & \texttt{false} & \texttt{top\_p}              & 0.8 \\
\texttt{top\_k}              & 20              & \texttt{temperature}         & 0 \\
\texttt{repetition\_penalty} & 1.0             & \texttt{presence\_penalty}  & 1.5 \\
\texttt{max-turns}            & 46              & \texttt{max-seq-lens}        & 48{,}000 \\
\bottomrule
\end{tabular}
\end{table}

\FloatBarrier
\section{More Results}
\label{app:more-results}

\subsection{Data Quality Analysis}
\label{app:data-quality-analysis}

The \revisiontext{ModalSearch-Pipeline} introduces image nodes into a knowledge graph and distinguishes stably retrievable images from retrieval-ambiguous images according to whether a textual description reliably identifies the visual content. To evaluate this judgment, we randomly sample 116 image nodes and use human annotations as the reference labels: 63 examples are stably retrievable (positive) and 53 are retrieval-ambiguous (negative). A positive example identifies essentially unique visual content, such as a specific moment in a historical event, a particular painting, an album cover, or a film poster, and image search with its description converges to that content. Negative examples have ambiguous descriptions or may lead to substantially different photographs, such as generalized events or ordinary scenes with different times, viewpoints, or weather conditions.

We compare Gemini-3.5-Flash, GPT-5.4, Gemini-2.5-Pro, and Claude-Sonnet-4.6 under the same rubric. Table~\ref{tab:uniqueness-judge} reports accuracy and Macro-F1, class-wise precision, recall, and F1 scores, and the corresponding confusion-matrix counts. Gemini-3.5-Flash achieves the highest accuracy (88.79\%) and Macro-F1 (88.33\%), with 100\% recall on the positive class. It retains all 63 positive examples while incorrectly retaining 13 of the 53 negative examples, demonstrating strong but imperfect agreement with human judgments. Balancing its leading classification accuracy with inference cost, we select Gemini-3.5-Flash as the referential-stability judge for the formal data-generation pipeline.

\begin{table}[!t]
\centering
\footnotesize
\setlength{\tabcolsep}{2.2pt}
\renewcommand{\arraystretch}{1.15}
\definecolor{tablebestblue}{RGB}{80,100,164}
\renewcommand{\textbf}[1]{{\bfseries\sffamily #1}}
\caption{Visual-retrievability judgments against human annotations on 116 examples: 63 stably retrievable (positive) and 53 retrieval-ambiguous (negative). Metric scores are percentages; TP, FP, FN, and TN are confusion-matrix counts. \textcolor{tablebestblue}{\textbf{Bold}} marks the best metric score in each column, including ties.}
\label{tab:uniqueness-judge}
\begin{tabular}{@{}l*{12}{c}@{}}
\toprule
\multirow{2}{*}{\textbf{Model}} & \multirow{2}{*}{\textbf{Accuracy}} & \multicolumn{3}{c}{\textbf{Positive class}} & \multicolumn{3}{c}{\textbf{Negative class}} & \multirow{2}{*}{\textbf{Macro-F1}} & \multicolumn{4}{c}{\textbf{Confusion matrix}} \\
\cmidrule(lr){3-5}\cmidrule(lr){6-8}\cmidrule(l){10-13}
& & \textbf{Precision} & \textbf{Recall} & \textbf{F1} & \textbf{Precision} & \textbf{Recall} & \textbf{F1} & & \textbf{TP} & \textbf{FP} & \textbf{FN} & \textbf{TN} \\
\midrule
Gemini-3.5-Flash & \textcolor{tablebestblue}{\textbf{88.79}} & 82.89 & \textcolor{tablebestblue}{\textbf{100.00}} & \textcolor{tablebestblue}{\textbf{90.65}} & \textcolor{tablebestblue}{\textbf{100.00}} & 75.47 & \textcolor{tablebestblue}{\textbf{86.02}} & \textcolor{tablebestblue}{\textbf{88.33}} & 63 & 13 & 0 & 40 \\
GPT-5.4 & 86.21 & \textcolor{tablebestblue}{\textbf{84.06}} & 92.06 & 87.88 & 89.36 & \textcolor{tablebestblue}{\textbf{79.25}} & 84.00 & 85.94 & 58 & 11 & 5 & 42 \\
Gemini-2.5-Pro & 83.62 & 76.83 & \textcolor{tablebestblue}{\textbf{100.00}} & 86.90 & \textcolor{tablebestblue}{\textbf{100.00}} & 64.15 & 78.16 & 82.53 & 63 & 19 & 0 & 34 \\
Claude-Sonnet-4.6 & 81.03 & 74.12 & \textcolor{tablebestblue}{\textbf{100.00}} & 85.14 & \textcolor{tablebestblue}{\textbf{100.00}} & 58.49 & 73.81 & 79.47 & 63 & 22 & 0 & 31 \\
\bottomrule
\end{tabular}
\end{table}

\FloatBarrier

\subsection{Inference Stability}
\label{app:inference-stability}

Because search tools may return different results across runs and agent outputs are stochastic for multi-turn trajectories, benchmark evaluations can exhibit non-negligible variation. To strengthen the reliability of the reported results and the conclusions drawn from them, we conduct two complementary stability analyses: repeated agent inference and repeated LLM judging on fixed agent outputs. Specifically, we run ModalSearcher-8B, Qwen3-VL-8B under the agentic-workflow setting, and Qwen3-VL-32B under the agentic-workflow setting five times on BrowseComp-VL, MMSearch, ModalSearch, and VisBrowse. We report the mean accuracy, variance, and 95\% confidence interval across the five inference runs in Table~\ref{tab:stability}. For both inference and repeated judging, variance is the sample variance of accuracy in squared percentage points, using an $n-1$ denominator. The 95\% confidence interval for the mean is $\bar a\pm t_{0.975,4}s/\sqrt{5}$, where $s$ is the sample standard deviation across the five evaluations. Although agent inference is stochastic, the observed variation is small relative to ModalSearcher's gains over the baselines, supporting the robustness of the main comparative conclusions.

In addition, for one fixed set of final summarized answers from the five inference runs above, we repeat the GPT-4o LLM-as-judge evaluation five times. The corresponding results are reported in the same table. Repeated judging generally exhibits lower variance than repeated agent inference, suggesting that judge-output randomness is not the dominant source of variation in these evaluations. This analysis measures repeatability rather than the correctness of the judge's decisions.

\begin{table}[H]
\centering
\small
\setlength{\tabcolsep}{4pt}
\renewcommand{\arraystretch}{1.1}
\caption{Inference and LLM-judge stability over five repeated evaluations. The first group of statistics reports repeated agent inference, while the second group reports repeated judging of one fixed inference result.}
\label{tab:stability}
\begin{adjustbox}{max width=\textwidth}
\begin{tabular}{@{}llcccccc@{}}
\toprule
\multirow{2}{*}{\textbf{Model}} & \multirow{2}{*}{\textbf{Benchmark}} & \multicolumn{3}{c}{\textbf{Inference}} & \multicolumn{3}{c}{\textbf{Judge}} \\
\cmidrule(lr){3-5}\cmidrule(l){6-8}
& & \textbf{Mean} & \textbf{Variance} & \textbf{95\% CI} & \textbf{Mean} & \textbf{Variance} & \textbf{95\% CI} \\
\midrule
\multirow{4}{*}{ModalSearcher-8B}
& BrowseComp-VL & 48.1\% & 0.7977 & [46.96\%, 49.18\%] & 48.9\% & 0.04397 & [48.66\%, 49.18\%] \\
& MMSearch & 70.3\% & 1.7444 & [68.63\%, 71.91\%] & 70.9\% & 0.08889 & [70.50\%, 71.24\%] \\
& ModalSearch & 45.8\% & 5.03333 & [43.01\%, 48.59\%] & 47.2\% & 0.20000 & [46.64\%, 47.76\%] \\
& VisBrowse & 33.4\% & 5.70708 & [30.41\%, 36.34\%] & 36.1\% & 0.35013 & [35.36\%, 36.83\%] \\
\midrule
\multirow{4}{*}{Qwen3-VL-8B (Agentic)}
& BrowseComp-VL & 31.6\% & 0.69095 & [30.55\%, 32.61\%] & 31.7\% & 0.01884 & [31.51\%, 31.85\%] \\
& MMSearch & 52.0\% & 0.16667 & [51.49\%, 52.51\%] & 51.8\% & 0.08889 & [51.43\%, 52.17\%] \\
& ModalSearch & 13.5\% & 3.42222 & [11.24\%, 15.83\%] & 16.4\% & 0.07778 & [16.05\%, 16.75\%] \\
& VisBrowse & 13.6\% & 2.80102 & [11.53\%, 15.69\%] & 15.1\% & 0.28010 & [14.49\%, 15.81\%] \\
\midrule
\multirow{4}{*}{Qwen3-VL-32B (Agentic)}
& BrowseComp-VL & 37.1\% & 0.31407 & [36.40\%, 37.79\%] & 36.8\% & 0.03141 & [36.62\%, 37.06\%] \\
& MMSearch & 58.3\% & 0.16667 & [57.83\%, 58.84\%] & 58.8\% & 0.03333 & [58.57\%, 59.03\%] \\
& ModalSearch & 27.7\% & 3.61111 & [25.31\%, 30.03\%] & 27.3\% & 0.07778 & [26.92\%, 27.61\%] \\
& VisBrowse & 21.9\% & 1.05038 & [20.62\%, 23.17\%] & 22.6\% & 0.24509 & [21.99\%, 23.22\%] \\
\bottomrule
\end{tabular}
\end{adjustbox}
\end{table}

\subsection{More Case Studies}
\label{app:more-case-studies}

\begin{figure}[!t]
    \centering
    \caption{A Search-required example generated by \revisiontext{ModalSearch-Pipeline}, with an initial image and additional visual evidence to be retrieved.}
    \label{fig:more_cases_1}
    \includegraphics[width=\textwidth]{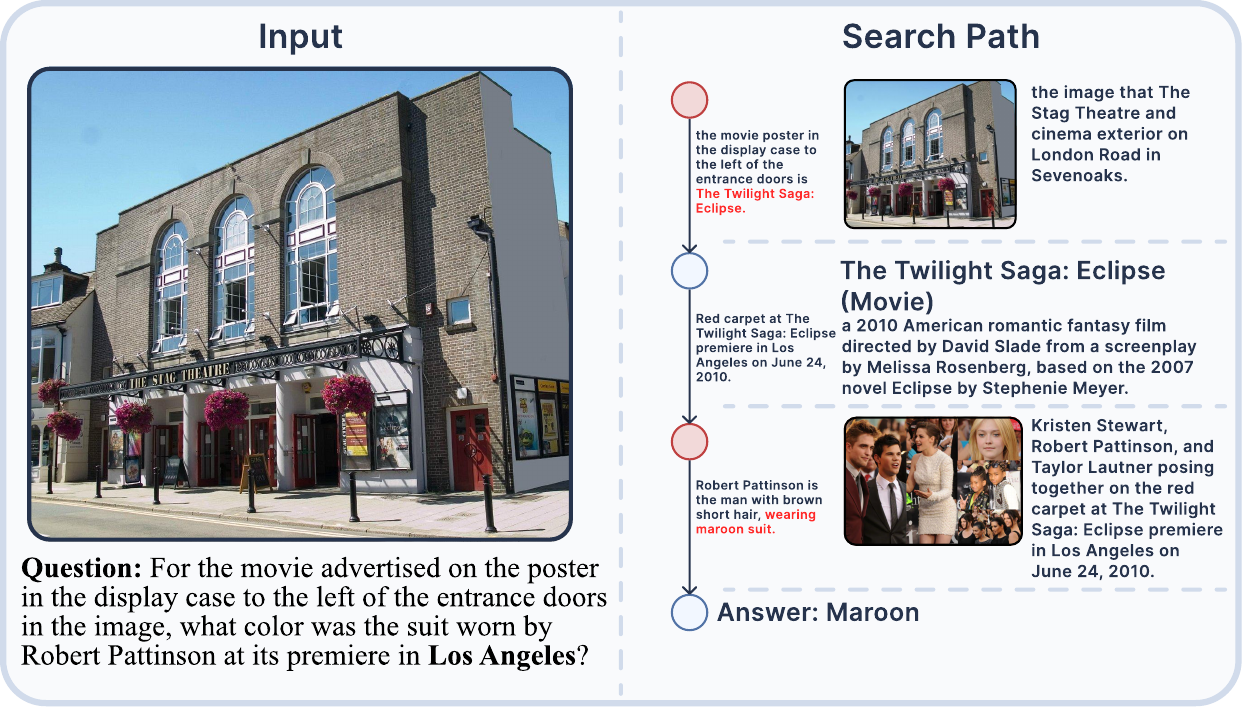}
\end{figure}

\begin{figure}[!htbp]
    \centering
    \caption{An Input-only example generated by \revisiontext{ModalSearch-Pipeline}, with visual evidence confined to the initial input.}
    \label{fig:more_cases_2}
    \includegraphics[width=\textwidth]{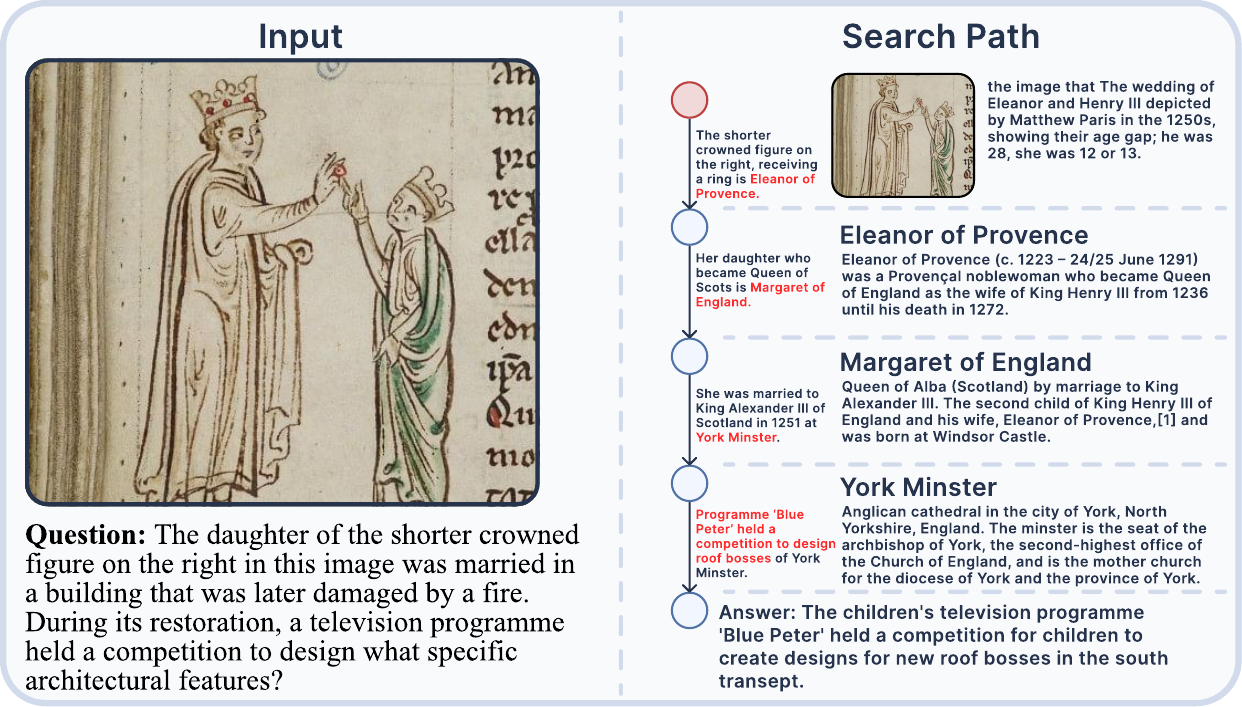}
\end{figure}

\begin{figure}[!htbp]
    \centering
    \caption{A Search-required example generated by \revisiontext{ModalSearch-Pipeline}, with interleaved textual and visual evidence.}
    \label{fig:more_cases_3}
    \includegraphics[width=\textwidth]{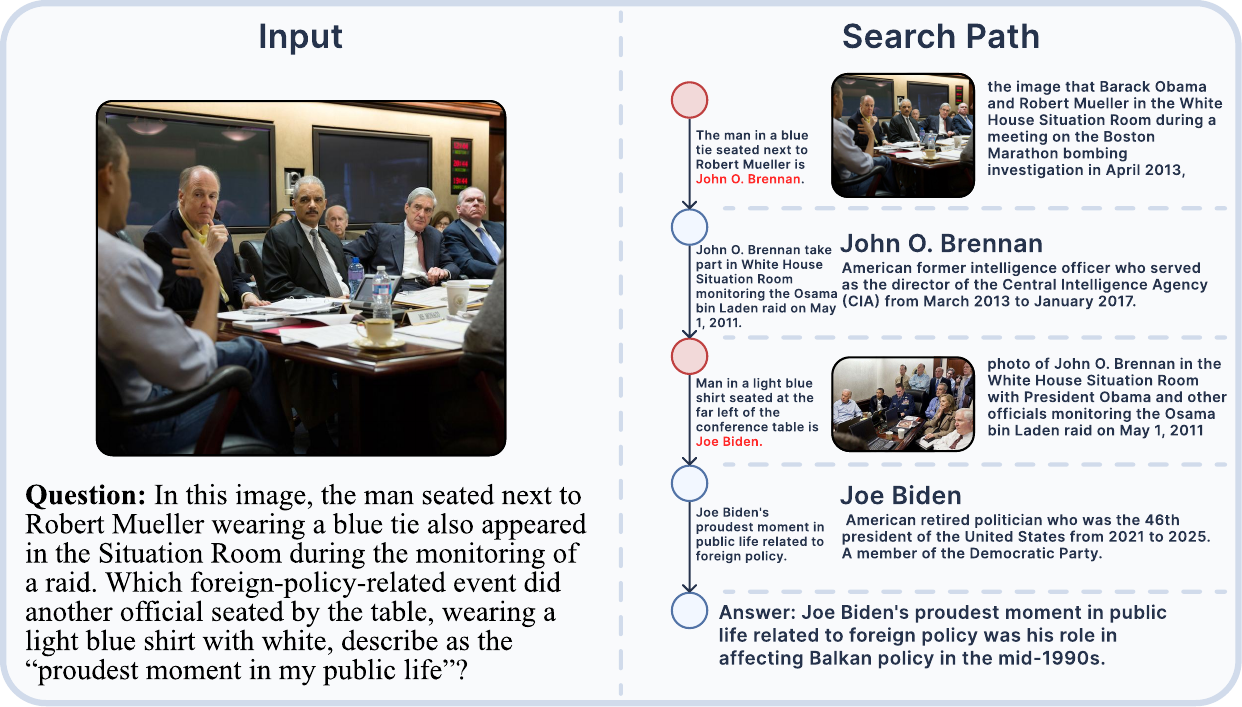}
\end{figure}

\begin{figure}[!htbp]
    \centering
    \caption{A Search-required example generated by \revisiontext{ModalSearch-Pipeline}, with no initial image and new visual evidence to be retrieved during reasoning.}
    \label{fig:more_cases_4}
    \includegraphics[width=\textwidth]{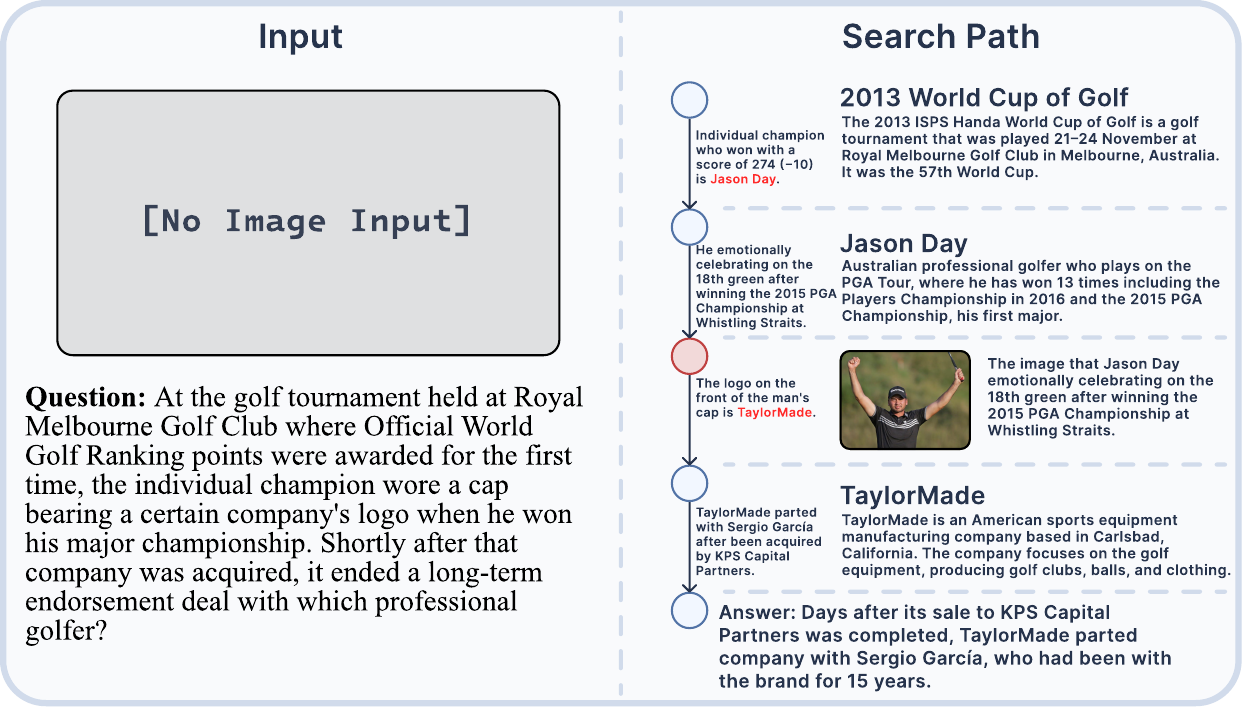}
\end{figure}

Figures~\ref{fig:more_cases_1}--\ref{fig:more_cases_4} present examples generated by \revisiontext{ModalSearch-Pipeline} across the Input-only and Search-required categories. Figure~\ref{fig:more_cases_2} uses only the initial visual evidence, \revisiontext{while the reference paths in Figures~\ref{fig:more_cases_1}, \ref{fig:more_cases_3}, and~\ref{fig:more_cases_4} include additional images to be retrieved during search.} By filtering out image nodes, the same pipeline can also generate purely textual data, outside this two-category multimodal taxonomy. This flexibility provides a unified generation process for unimodal and multimodal multi-hop search problems and offers a scalable recipe for training large-scale Web search agents. More broadly, the design of an interleaved multimodal knowledge graph provides a basis for extending the pipeline to additional modalities, such as audio and video.

As a representative example, Figure~\ref{fig:more_cases_1} shows a Search-required instance that also includes an initial image. During data construction, the sampled path starts from a street-view image of a theater and connects to the movie poster displayed beside its entrance, then to the film \emph{The Twilight Saga: Eclipse}, its premiere in Los Angeles, and a related red-carpet image. The final answer is the color of the suit worn by Robert Pattinson. \revisiontext{The backward question-construction process hides the movie title and direct identifiers of the target image that would expose reasoning shortcuts, while retaining Robert Pattinson and Los Angeles as necessary disambiguating clues in the final question.} The resulting ideal evidence path therefore requires identifying the film from the initial visual clue, retrieving information about its premiere and the associated visual evidence, and finally determining that the answer is \emph{maroon}. This path is a reference evidence chain defined during data construction, rather than a fixed trajectory that the agent must follow at inference time.

\begin{figure}[!htbp]
    \centering
    \caption{A real ModalSearcher-8B inference trajectory on the Search-required category of ModalSearch-Bench, involving reverse-image search, text retrieval, and visual verification of a historical painting.}
    \label{fig:agent_case_1}
    \includegraphics[width=0.90\textwidth]{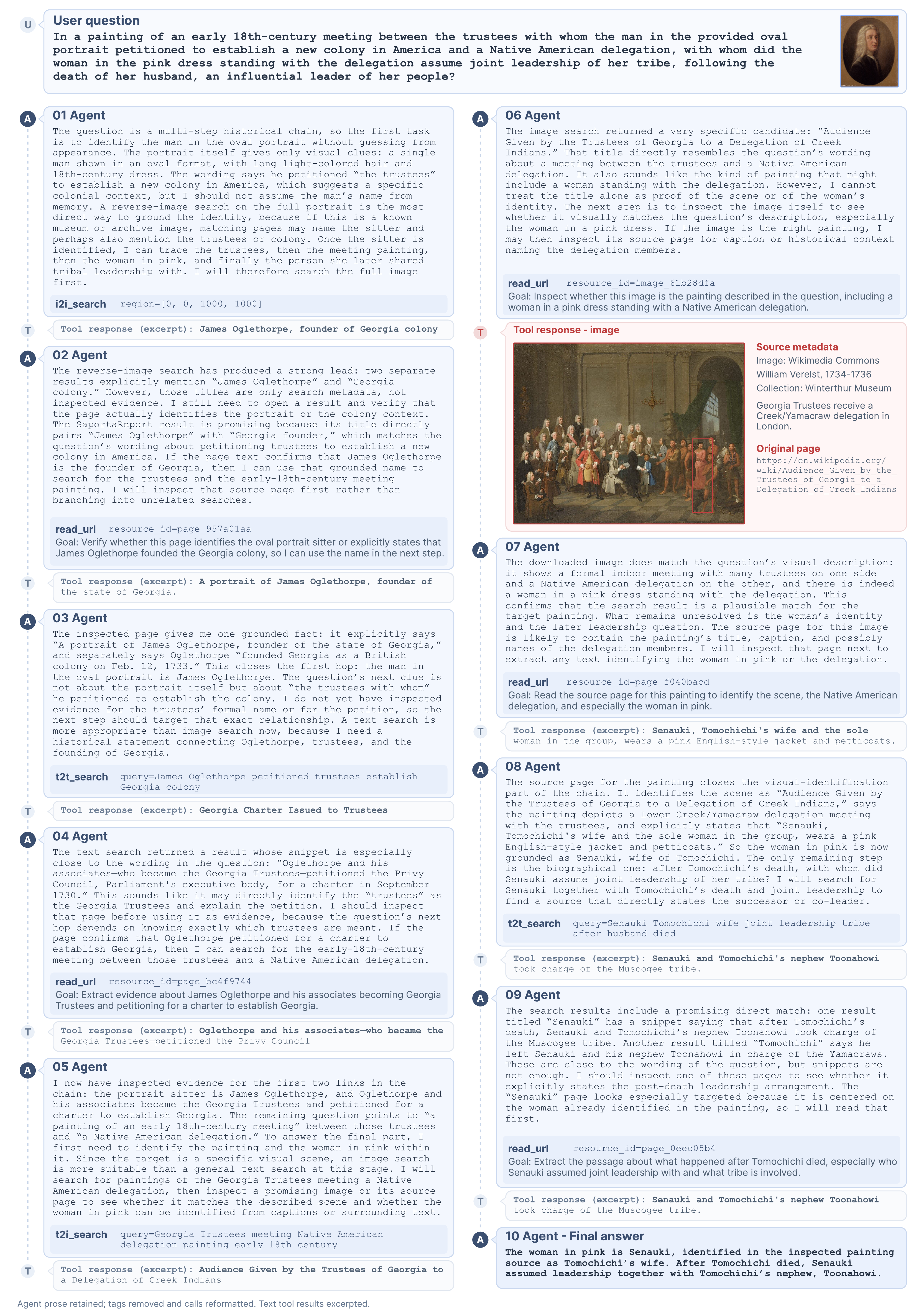}
\end{figure}

\begin{figure}[!htbp]
    \centering
    \caption{A real ModalSearcher-8B inference trajectory on the Search-required category of ModalSearch-Bench, combining textual entity tracing with retrieval and inspection of new visual evidence.}
    \label{fig:agent_case_2}
    \includegraphics[width=0.90\textwidth]{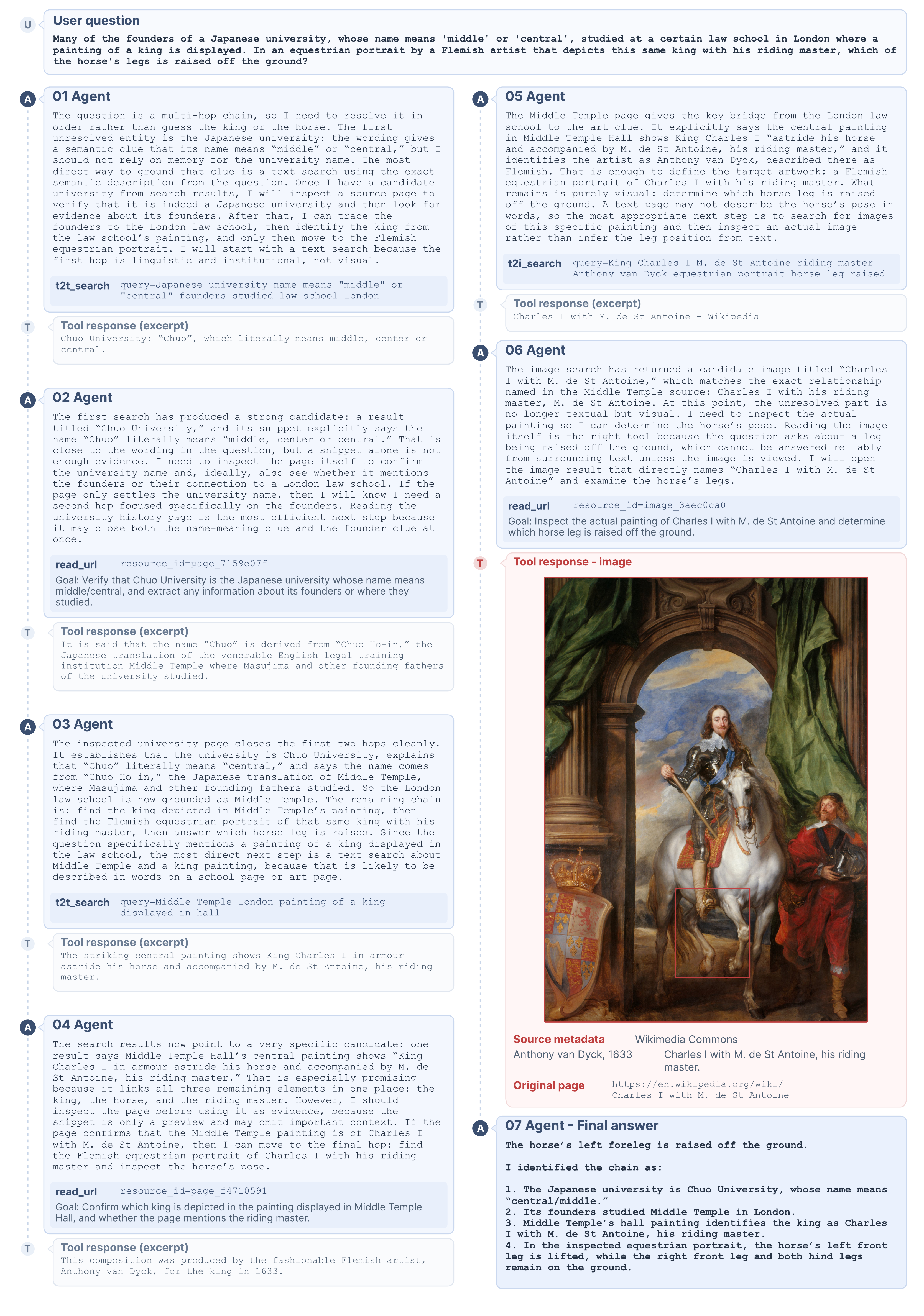}
\end{figure}

We additionally show two real inference trajectories of ModalSearcher-8B on ModalSearch-Bench in Figures~\ref{fig:agent_case_1} and~\ref{fig:agent_case_2}. For Figure~\ref{fig:agent_case_1}, the question asks the model to identify the person in an oval portrait, trace the painting depicting a meeting between the Georgia Trustees and a Native American delegation, and determine with whom the woman in pink later shared tribal leadership after her husband's death. The agent first performs a reverse-image search and identifies the sitter as \emph{James Oglethorpe}. It then uses text search to verify his connection to the Georgia Trustees and to locate the early-eighteenth-century painting depicting the meeting. After opening the candidate image, the agent visually confirms the woman in pink and follows the source page to identify her as \emph{Senauki}. It finally searches for Senauki's biography and finds that, after \emph{Tomochichi}'s death, she shared leadership with his nephew \emph{Toonahowi}. This trajectory requires the agent to alternate between visual identification, textual retrieval, and image-grounded verification.

For Figure~\ref{fig:agent_case_2}, the question asks the model to identify a Japanese university from semantic clues, trace its founders to the London law institution whose collection contains a painting of a king, and then determine the pose of the horse in a related equestrian portrait. The agent first verifies the university as Chuo University and traces the founders' connection to Middle Temple in London. It then identifies the painting as an equestrian portrait of \emph{King Charles I} with his riding master \emph{M. de St. Antoine}, painted by \emph{Anthony van Dyck}. After retrieving the corresponding image, the agent inspects the visual details and determines that the horse's left foreleg is raised off the ground. Together, these two cases illustrate how ModalSearcher-8B switches modalities at appropriate points and integrates visual and textual evidence when solving interleaved multimodal, multi-hop search problems. Unlike the preceding constructed examples, these figures show the model's actual inference trajectories on ModalSearch-Bench.

\FloatBarrier
\section{Inference and Evaluation Prompts}
\label{app:used-prompts}

\begin{figure}[!ht]
    \centering
    \caption{System prompt used by the multimodal research agent during inference.}
    \label{fig:prompt_inference}
    \includegraphics[width=0.60\textwidth]{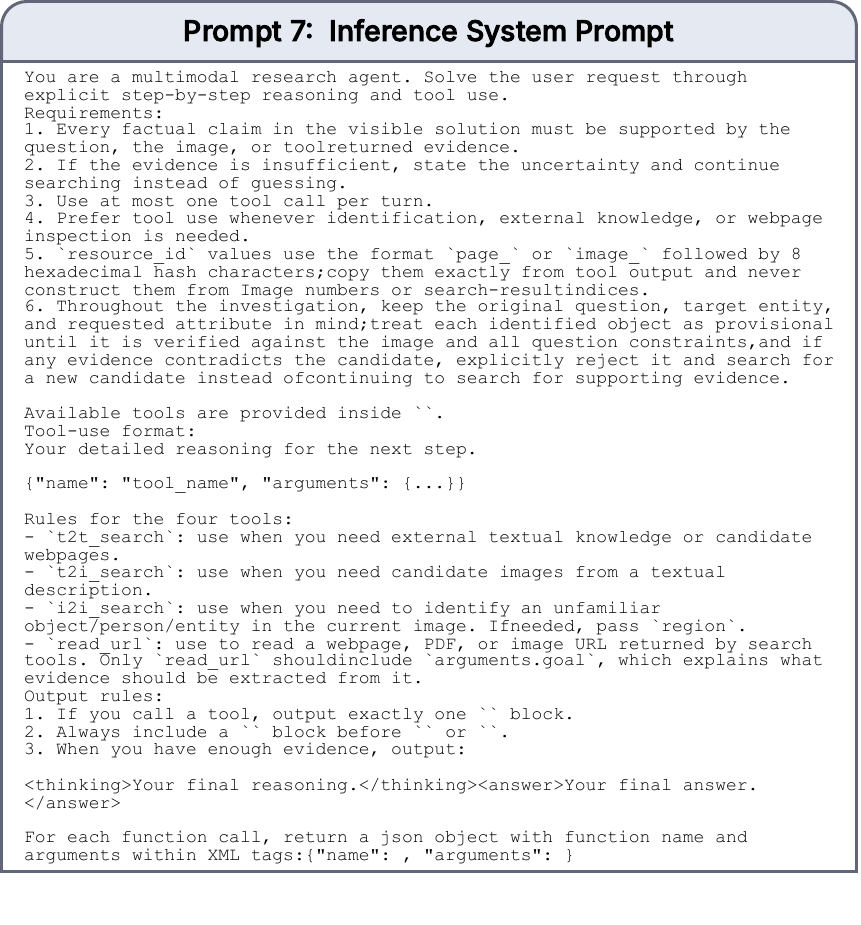}
    \vspace{0.2em}
    \caption{Prompt used by GPT-4o for LLM-as-judge evaluation.}
    \label{fig:prompt_evaluation}
    \includegraphics[width=0.60\textwidth]{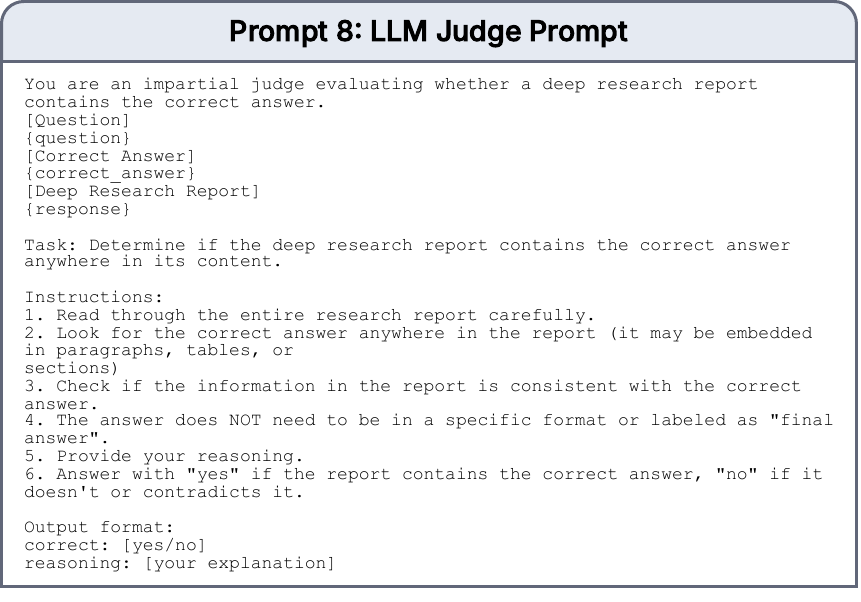}
\end{figure}

This section presents the agent inference and answer-evaluation prompts. Figure~\ref{fig:prompt_inference} shows the system prompt used by the multimodal research agent during inference. Figure~\ref{fig:prompt_evaluation} shows the GPT-4o LLM-as-judge prompt, which is aligned with OpenSearch-VL to ensure consistency with its evaluation protocol.

\FloatBarrier

\FloatBarrier

\end{document}